%% file: main.tex
\documentclass[10pt,twocolumn,letterpaper]{article}

\usepackage[pagenumbers]{cvpr} 

\input{preamble}

\definecolor{cvprblue}{rgb}{0.21,0.49,0.74}
\usepackage[pagebackref,breaklinks,colorlinks,citecolor=cvprblue]{hyperref}

\title{Three-view Focal Length Recovery From Homographies}

\author{Yaqing Ding$^1$ \quad Viktor Kocur$^2$ \quad Zuzana Berger Haladová$^2$ \quad Qianliang Wu$^3$ \\
Shen Cai$^4$ \quad Jian Yang$^3$ \quad Zuzana Kukelova$^1$\vspace{0.25em}
\\
$^{1}$ Visual Recognition Group,
Faculty of Electrical Engineering, Czech Technical University in Prague \\
{\tt\small \{yaqing.ding, kukelzuz\}@fel.cvut.cz} \\
$^{2}$ Faculty of Mathematics, Physics and Informatics, Comenius University in Bratislava \\
{\tt\small \{viktor.kocur, haladova\}@fmph.uniba.sk} \\
$^{3}$ PCA Lab, Nanjing University of Science and Technology, Nanjing, China\\
{\tt\small \{wuqiangliang, csjyang\}@njust.edu.cn} \\
$^{4}$ Visual and Geometric Perception Lab, Donghua University\\
{\tt\small   hammer\_cai@163.com}\\
}

\begin{document}
\maketitle
\input{sec/0_abstract}   
\input{sec/1_intro}

\input{sec/2_related}

\input{sec/3_problem}

\input{sec/4_approach}
\input{sec/5_solvers}

\input{sec/6_synthetic}

\input{sec/7_real}
\input{sec/8_conclusion}
{
    \small
    \bibliographystyle{ieeenat_fullname}
    \bibliography{main}
}

\input{sec/X_suppl}

\end{document}

%% file: preamble.tex
\usepackage{graphicx}
\usepackage[table,dvipsnames]{xcolor}
\usepackage{booktabs}
\usepackage{tikz}
\usepackage{siunitx}
\usepackage{verbatim}
\usepackage{booktabs}
\usepackage{multirow}
\usepackage{makecell}
\usepackage{pifont} 
\usepackage{soul}
\usepackage{rotating}
\usepackage{overpic}
\usepackage{pdflscape}

\definecolor{mygray}{gray}{.92}
\newsavebox\CBox

\newcommand{\M}[1]{\mathbf{#1}}
\definecolor{wincolor}{rgb}{0.95, 0.2, 0.2}

\DeclareMathSymbol{\shortminus}{\mathbin}{AMSa}{"39}

\usepackage{pgfplots}
\pgfplotsset{width=10cm,compat=1.9}
\definecolor{Seaborn1}{HTML}{1f77b4}
\definecolor{Seaborn2}{HTML}{ff7f0e}
\definecolor{Seaborn3}{HTML}{2ca02c}
\definecolor{Seaborn4}{HTML}{d62728}
\definecolor{Seaborn5}{HTML}{9467bd}
\definecolor{Seaborn6}{HTML}{8c564b}
\definecolor{Seaborn7}{HTML}{e377c2}
\definecolor{Seaborn8}{HTML}{7f7f7f}
\definecolor{Seaborn9}{HTML}{bcbd22}

\pgfplotsset{
compat=1.9,
legend image code/.code={
\draw[mark repeat=4,mark phase=1]
plot coordinates {
(0cm,0.02cm)
(0.1cm,0.02cm)        
(0.2cm,0.02cm)
(0.3cm,0.02cm)      
};%
}
}

\newcommand{\case}[1]{Case \MakeUppercase{\romannumeral#1}}

\newcommand{\h}[1]{$\M H_{#1}$}
\newcommand{\hr}[1]{$\M H_{#1}+\M P3\M P$}
\newcommand{\hrf}[1]{$\M H_{#1}+\M P4\M Pf$}
\newcommand{\fEf}{$f\M{E}f$}
\newcommand{\fEfr}{$f\M{E}f + \M P3\M P$}
\newcommand{\fEfrf}{$f\M{E}f + \M P4\M Pf$}
\newcommand{\fEfp}{$f\M{E}f + \M P\M P$}
\newcommand{\fEfpr}{$f\M{E}f + \M P\M P + \M P3\M P$}
\newcommand{\fEfprf}{$f\M{E}f + \M P\M P + \M P4\M Pf$}
\newcommand{\Ef}{$\M{E}f$}
\newcommand{\Efr}{$\M{E}f + \M P3\M P$}
\newcommand{\Efrf}{$\M{E}f + \M P4\M Pf$}

\newcommand{\pnp}[1]{$\M{P}#1\M{P}$}
\newcommand{\pnpf}{$\M P4\M Pf$}

\newcommand{\PAR}[1]{\vskip4pt \noindent{\bf #1~}}

%% file: sec/0_abstract.tex
\begin{abstract}
\noindent
In this paper, we propose a novel approach for recovering focal lengths from three-view homographies. By examining the consistency of normal vectors between two homographies, we derive new explicit constraints between the focal lengths and homographies using an elimination technique. We demonstrate that three-view homographies provide two additional constraints, enabling the recovery of one or two focal lengths. We discuss four possible cases, including three cameras having an unknown equal focal length, three cameras having two different unknown focal lengths, three cameras where one focal length is known, and the other two cameras have equal or different unknown focal lengths. All the problems can be converted into solving polynomials in one or two unknowns, which can be efficiently solved using Sturm sequence or hidden variable technique. Evaluation using both synthetic and real data shows that the proposed solvers are both faster and more accurate than methods relying on existing two-view solvers. The code and data are available on \url{https://github.com/kocurvik/hf}.


\end{abstract}

%% file: sec/1_intro.tex
\section{Introduction}
\label{sec:intro}

Estimating relative camera motion from multiple views using point correspondences is a classical problem in computer vision. Efficient solutions for various camera configurations have been well-studied in the literature~\cite{nister2004efficient,hartley2012efficient,kukelova2012polynomial,stewenius2005minimal,li2006simple,brown2007minimal,hartley1992estimation,hartley2003multiple,hartley1997defense,nister2006Four}. For example, with two fully calibrated cameras, the relative camera pose can be efficiently determined using 5-point algorithms~\cite{nister2004efficient,hartley2012efficient,kukelova2012polynomial}. 
In contrast to the two view case, the calibrated three-view relative pose problem is much more challenging. 
This problem can be solved using four triplets of point correspondences~\cite{nister2006Four,quan2006Results},\footnote{Note, that configuration of four points in three views generates an over-constrained problem. In this case, we have 12 constraint for 11 degrees of freedom (DOF).  A minimal solution would need to drop one constraint, \eg, 
by considering only a line passing through one of the points in the third view~\cite{hruby2022Learning} or by considering a ``half" point correspondence.} however it results in a very complex system of polynomial equations. 
Thus, the existing solutions~\cite{nister2006Four,hruby2022Learning} to this problem are only approximate and can often fail, \ie, the returned solution can be arbitrarily far from the geometrically correct solution.


\begin{figure}[t]
 \begin{overpic}[width=0.82\columnwidth]{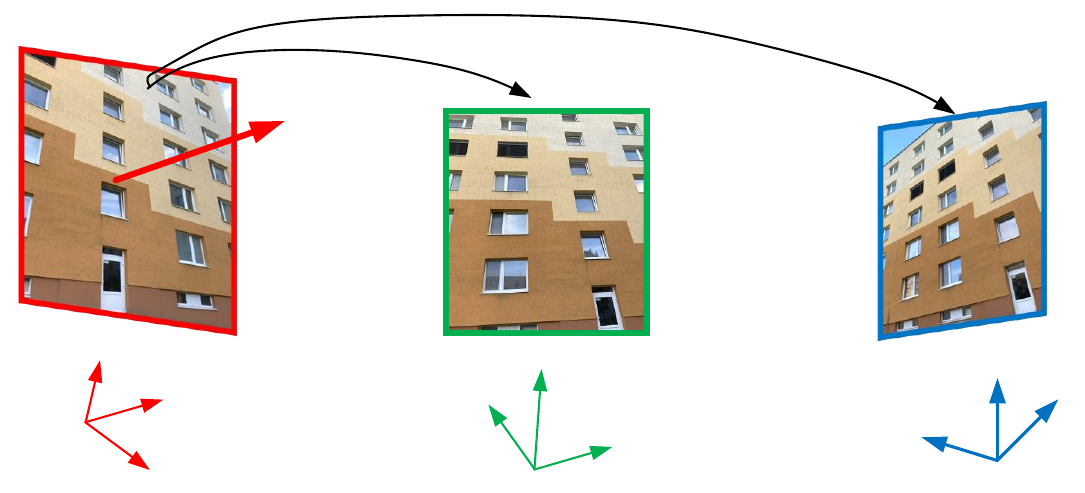}

    \put(13,4) {\small $[\M{I}~\M{0}]$}
    \put(51,6) {\small $[\M{R}_2~\M{t}_2]$}
    \put(78,7) {\small $[\M{R}_3~\M{t}_3]$}
    \put(51,37) {\small $\M H_2$}
    \put(88,37) {\small $\M H_3$}
    \put(25,31) {\small $\M n$}
    \end{overpic}
    \caption{Three cameras view the same plane, defining two homographies $\M H_2$, $\M H_3$. The two homographies have the same reference image, which should correspond to the same normal vector $\M n$.}\label{fig:teaser}
\end{figure}
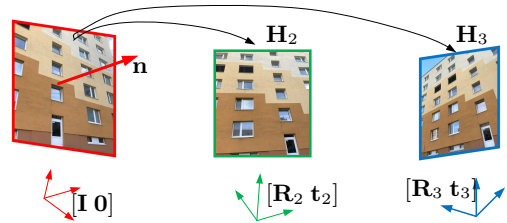

An important scenario arises when the focal length is the only unknown in the camera intrinsic parameters, referred to as the partially calibrated case. This setup is practical since most modern cameras have zero skew and a centered principal point. When two cameras share an equal yet unknown focal length, their relative motion and the common focal length can be estimated using six point correspondences~\cite{stewenius2005minimal,hartley2012efficient,kukelova2012polynomial,kukelova2017clever}. Similarly, if one camera is fully calibrated and the other has an unknown focal length, six point correspondences are also required~\cite{bujnak20093d,kukelova2012polynomial,kukelova2017clever}. For two cameras with different, unknown focal lengths, at least seven point correspondences are necessary to recover both the relative motion and the focal lengths~\cite{hartley1992estimation,bougnoux1998projective}.

Three-view focal length problems are significantly more complex, often involving polynomial systems with hundreds of solutions. For example, when the three cameras share the same focal length, there can be up to 668 possible solutions~\cite{chien2022gpu,ding2023minimal,cin2024minimal}. Such systems can  be solved using homotopy continuation methods~\cite{sommese2005numerical, hauenstein2018adaptive}. 
The running times of the CPU variants of the solvers for four points in three cameras with unknown shared focal length range from $250ms$ to $1456ms$. 
Efficient GPU implementations are much faster; however, with runtimes $16.7ms$ to $154ms$ they are still too slow for practical applications.

Scenes with planar surfaces, such as floors, walls, doors, facades, and other common structures, are prevalent in man-made environments. 
When points are coplanar, homography-based algorithms require four point correspondences~\cite{hartley2003multiple} to estimate the relative pose of two cameras. Unfortunately, any attempt to recover intrinsic parameters from two views of a planar surface (using only point correspondences without additional priors) is futile, as stated in the following theorem~\cite{maybank1993theory,nister2004efficient}: "For any choice of intrinsic parameters, any homography can be realized between two views by positioning the views and a plane in a particular way." However, it has been shown that focal lengths can be recovered from three-view homographies~\cite{malis2002camera,heikkila2017using}, \ie three cameras observing a planar surface (see illustration in Figure~\ref{fig:teaser}).

In this paper, we propose novel solutions to the problem of estimating the focal lengths from three-view homographies.
We consider four possible cases: (i) three cameras having an unknown equal focal length, (ii) three cameras having two different unknown focal lengths, (iii) three cameras with one known focal length and two unknown equal focal lengths, and (iv) three cameras with one known focal length and two unknown different focal lengths (see Table~\ref{tab:cases}). 
We propose novel formulations to these problems and discuss the number of possible solutions for each case. 
The main contributions of the paper include: 
\begin{itemize}
	\item We solve the three-view focal length problems using four coplanar points. We use the property that four coplanar points in three views define two independent homographies, which should share the same plane normal vector. 
        \item Based on the normal vector consistence, we derive new explicit constraints on the focal lengths and the homographies. We provide a detailed problem formulation for the above-mentioned cases, which are then converted into solving polynomial systems in only one or two unknowns.
	\item This allows us to develop solvers using efficient Sturm sequences or hidden variable technique.
    The proposed solvers are significantly more efficient than the existing solvers to this problem~\cite{heikkila2017using}.
    Moreover, in extensive synthetic and real experiments, we show an improvement in accuracy over the state-of-the-art two-view solvers.
    \item We propose a new dataset consisting of 6 scenes (four indoor and 2 outdoor) containing 1870 images captured with 14 different cameras. 
    Ground truth focal lengths are estimated using the standard calibration method~\cite{zhang2000calibration}.    
   We will make both the dataset and the 
   code publicly available.
    \item To our knowledge, we are the first to extensively evaluate focal length self-calibration from three views of planar scenes on large amounts of synthetic and real scenes, and compare the solutions with different two-view and three-view baselines for cameras observing general scenes.
\end{itemize}



%% file: sec/2_related.tex
\section{Related Work}
\label{sec:related}

Homography estimation is a well-studied problem in the literature and can be solved using the 4-point algorithm, which involves solving a system of eight homogeneous linear equations~\cite{hartley2003multiple}. However, with only point correspondences, two-view homography does not provide additional constraints on the focal length, making the two-view homography not suitable for self-calibration. To recover the intrinsic parameters from the two-view homography, we 
need additional priors. In~\cite{brown2007minimal}, under pure rotation assumption, the authors propose two minimal solutions for the two-view homography-based focal length estimation with one or two focal length parameters. In~\cite{ding2020homography}, the authors use sensor-fusion,~\ie, combining camera with gravity from IMU (Inertial measurement unit), to reduce the DOF of the motion parameters. In general, the gravity prior can provide up to two constraints on the focal length. 

Focal length recovery from 
homographies in more views
was first discussed in~\cite{malis2002camera}. Malis~\etal compute homographies among multiple views, iteratively testing various focal lengths to find the one that minimizes a cost function. However, this method is 
time-consuming and sensitive to noise.

The work most closely related to ours is by Heikkilä~\cite{heikkila2017using}, where the constraints on the focal length and the normal vector are derived, and non-iterative solutions to the problem of recovering focal lengths from three-view homographies is proposed. For three cameras with the same unknown focal length, 
~\cite{heikkila2017using}
solves three polynomials in three unknowns. However, these polynomials have very high degrees, making them challenging to solve efficiently. Heikkilä first applies SVD 
to a $82\times 82$ matrix, followed by Gaussian-Jordan elimination of a $82\times 164$ matrix. Finally, the solver finds the solutions by computing the eigenvalues of a $82\times 82$ matrix. For three cameras with two different focal lengths, Heikkilä’s method needs to solve four polynomials in four unknowns by performing Gaussian-Jordan elimination of a $176\times 352$ matrix and computing the eigenvalues of a $176\times 176$ matrix. 

In contrast to complex solutions presented in~\cite{heikkila2017using}, we propose very efficient solutions to the problem of recovering focal lengths from three-view homographies.
For 
cameras with the same focal length, our novel solver only requires computing the roots of a univariate polynomial of degree 9, which can be efficiently solved using Sturm sequences. For 
cameras with two different focal lengths, we compute solutions from the eigenvalues of a $18\times 18$ matrix. 

\begin{table*}[ht]
\begin{center}
\resizebox{\linewidth}{!}
        { 
  \begin{tabular}{lccccccccccccccc}
    \toprule
    &&&\multicolumn{3}{c}{Focal length}  && \multicolumn{2}{c}{No. of Solutions} && \multicolumn{2}{c}{Eigenvalue} && \multicolumn{2}{c}{Time ($\mu s$)}  \\ \cmidrule{4-6} \cmidrule{8-9}  \cmidrule{11-12}  \cmidrule{14-15} 
    Problems& Proposed Solvers & & View 1 & View 2 & View 3 && Ours & Heikkilä~\cite{heikkila2017using}&& Ours & Heikkilä~\cite{heikkila2017using}  && Ours & Heikkilä~\cite{heikkila2017using} \\
      \midrule
   Case \MakeUppercase{\romannumeral1} & \h{fff} &&$f$ & $f$ & $f$  && 9 & 70 && Sturm & $82\times 82$ && 17.3 & 1404 \\
   Case \MakeUppercase{\romannumeral2} & \h{ff} && Known & $f$ & $f$ &&  6 & - && Sturm & -  && 19.4 & - \\
     \midrule
    Case \MakeUppercase{\romannumeral3} & \h{f\rho\rho} &&$f$ & $\rho$ & $\rho$  && 17 & 152 && $18\times 18$ & $176\times 176$  && 200 & 5486 \\
  Case \MakeUppercase{\romannumeral4} & \h{f\rho} && Known & $f$ & $\rho$  && 9 & - && $12\times 12$ & -  && 106  & - \\
    \bottomrule
  \end{tabular} 
  }
\end{center}
  \caption{Four possible cases for three views. \romannumeral1) Equal and unknown focal length for three cameras; \romannumeral2) Known focal length of reference camera, equal and unknown focal length for target cameras; \romannumeral3) Focal lengths of reference camera and two target cameras are different; \romannumeral4) Known focal length of reference camera, different and unknown focal lengths for target cameras.}
  \label{tab:cases}   
\end{table*}

For general scenes, focal lenghts can be recovered together with the epipolar geometry.
For two cameras with equal unknown focal lengths, the minimal 6-point solvers~\cite{stewenius2005minimal,bujnak20093d,hartley2012efficient,kukelova2012polynomial,kukelova2017clever} use rank-2 and trace constraints on the essential matrix. The 5-point solver~\cite{torii2011six} solves the plane+parallax scenario for cameras with unknown focal length using four coplanar points and one off-plane point.
This solver can be used within the DEGENSAC framework~\cite{chum2005degensac} to complement the six point solvers in order to improve performance for scenes that are not completely planar, but contain a dominant plane.
For the single unknown focal length problem, the two-view geometry can be solved using the single-side 6-point solver~\cite{bujnak20093d,kukelova2017clever}. For the relative pose problem with different and unknown focal lengths, the solution computes the  fundamental matrix using the 7-point solver~\cite{hartley2003multiple}, followed by the extraction of the focal lengths from 
this
matrix~\cite{bougnoux1998projective, kocur2024robust}.



As discussed above, three-view focal length problems for general scenes are significantly more complex and their homotopy continuation solutions~\cite{chien2022gpu,ding2023minimal,cin2024minimal} are
impractical.

%% file: sec/3_problem.tex
\section{Problem Statement}
\label{sec:Problem}
In this section we introduce the three-view focal length problem using coplanar points along with the used notation. We consider a set of 3D points $\{\M X_i\}$ which lie on a plane defined by
\begin{equation}
	\M n^\top\M X_{i}=d, \label{eq:plane}
\end{equation}
where $\M n=[n_x,n_y,n_z]$ is the unit normal of the plane and $d$ is the distance of the plane to origin.

The points are observed by three cameras such that $\M X_i$ is projected to a 2D point $\M m_{i,j}$ by the $j$-th camera $\M K_j[\M R_j\ |\ \M t_j]$. In many practical scenarios, it is often reasonable to assume that the cameras have square-shaped pixels, and the principal point coincides with the image center~\cite{hartley2012efficient}. This leaves only the focal point unknown and thus $\M K_j = {\rm diag}(f_j,f_j,1)$. 

Without loss of generality, we set the coordinate system such that $\M R_1 = \M I$ and $\M t_1 = \M 0$. We use $\M X_{i,j}$ to denote the points $\{\M X_i\}$ expressed in the coordinate system of the $j$-th camera giving us
\begin{align}
	 \M X_{i,j} &= \M R_j \M X_{i} +\M t_j, j=2,3. \label{eq:camera_coords}
\end{align}
We can express the 3D point $\M X_{i, j}$ using the 2D projections $\M m_{i,j}$ as
\begin{equation}
    \M X_{i,j} = \lambda_{i, j} \M K_j^{-1}\M m_{i, j},
    \label{eq:lambda}
\end{equation}
where $\lambda_{i, j}$ is the depth of the point. Substituting \eqref{eq:plane} into \eqref{eq:camera_coords} we obtain
\begin{equation}    
	 \M X_{i,j} = \M R_j \M X_{i} + \frac{\M t_j}{d}\M n^\top \M X_{i} = \M H_j \M X_i,
\label{eq:v03}
\end{equation}
where $\M H_j = \M R_j+\frac{\M t_j}{d}\M n^\top$ is the Euclidean homography matrix. Further substituting \eqref{eq:lambda} into \eqref{eq:v03} and expressing $\M X_i$ in~\eqref{eq:v03} in the coordinate system of the first camera, we obtain
\begin{equation}
	\lambda_{i,j} \M K_j^{-1} \M m_{i,j} =\lambda_{i, 1} \M H_j \M K_1^{-1}  \M m_{i, 1}, 
\label{eq:v05}
\end{equation}
which can be reformulated as
\begin{align}
\begin{split}
  &[\M m_{i,j}]_\times\M G_j \M m_{i, 1} = \M 0,\\
   & {\rm with}\ \M G_j \sim \M K_j \M H_j \M K_1^{-1},
   \end{split}
\label{eq:v06}
\end{align}
where $\sim$ indicates equality up to a scale factor and $\M G_j$ represents homography between the first and the $j^{th}$ camera. 

By using point correspondences between two images we can use equations~\eqref{eq:v06} to obtain 2D homographies $\M G_j$. Our aim is to estimate the unknown focal lengths $f_j$ from $\M G_2$ and $\M G_3$. There are several possible configurations of three cameras 
based on combinations of known and unknown equal or different focal lengths. 
In Section~\ref{sec:solvers} we derive solvers for four such camera configurations denoted as \case{1}-\MakeUppercase{\romannumeral4}, which we list in Table~\ref{tab:cases}. 

We derive the new solvers by utilizing a key observation that both $\M H_2$ and $\M H_3$ are related to the normal vector $\M n$. In the next section, we use this to show that $\M G_2$ and $\M G_3$ can provide two constraints on the intrinsic parameters, which can be used to recover the focal lengths.

%% file: sec/4_approach.tex
\section{Our Approach}\label{sec:approach}

Given two 2D homographies $\M G_j$, we have
\begin{align}
\begin{split}
\M H_j \sim \M K_j^{-1} \M G_j \M K_1. \\
 \end{split}
\label{eq:07}
\end{align}
Thus for known  $\M G_j$,  $\M H_j$ are polynomial matrices in the focal length parameters. To solve for the focal lengths, we need to find the constraints on $\M H_j$. Based on the formulation of the Euclidean homography~\eqref{eq:v03}, we know  that~\cite{hartley2003multiple}
\begin{align}
	{\M E}_{j} = [\M t_j]_\times \M H_j = [\M t_j]_\times(\M R_j+\frac{\M t_j}{d}\M n^\top) = [\M t_j]_\times \M R_j,
\label{eq:08}
\end{align}
where $\M E_{j}$ is the essential matrix corresponding to $\M H_j$. Let's consider $\M H_j^\top$, where
\begin{align}
	 \M H_j^\top &= \M R_j^\top+\frac{\M n}{d}\M t_j^\top.
\label{eq:09}
\end{align}
It can be easily shown that the essential matrix $\tilde{\M E}_{j}$ corresponding to $\M H_j^\top$ is given by 
\begin{align}
    \tilde{\M E}_{j} = [\M n]_\times \M H_j^\top = [\M n]_\times(\M R_j^\top+\frac{\M n}{d}\M t_j^\top) = [\M n]_\times \M R_j^\top.
\label{eq:10}
\end{align}
Thus the essential matrices derived from different $\M H_j^\top$ are related by the same normal vector $\M n$ in the reference camera coordinate. As shown in~\cite{stefanovic1973relative,nister2004efficient}, a valid essential matrix should satisfy the singular and trace constraints
\begin{align}
\begin{split}
&\det(\tilde{\M E}_{j})=0, \\
 \tilde{\M E}_{j} \tilde{\M E}_{j}^\top \tilde{\M E}_{j} - &\frac{1}{2} {\rm trace}(\tilde{\M E}_{j} \tilde{\M E}_{j}^\top) \tilde{\M E}_{j} = \M 0.
 \end{split}
\label{eq:11}
\end{align}
In our case, we omit the zero determinant constraint since $\tilde{\M E}_{j}$ is already singular by construction~\eqref{eq:10}. 


Substituting~\eqref{eq:07} into the trace constraints~\eqref{eq:11}  gives us nine equations per homography, \ie, 18 equations for two homography matrices $\M{H}_2$ and $\M{H}_3$. 
Only two from the nine equations are algebraically independent, \ie, one homography matrix provides two constraints on the focal lengths and the normal vector parameters. For the case of equal unknown focal length, we have only 3 unknowns $(f,n_x,n_y)$ (since~\eqref{eq:11} is homogenous in $\M n$, we can let $n_z=1$) and thus an over-constrained problem. The problem can be solved using only one of the nine trace constraint equations for $\tilde{\M E}_{2}$. These equations can be solved using the Gr\"{o}bner basis method~\cite{larsson2017efficient,larsson2018beyond}
The final solver performs Gauss-Jordan elimination of a $118\times141$ matrix and extracts solutions from the eigenvectors of a $23\times 23$ matrix. However, this solver is not efficient enough for real-time applications.

To derive more efficient solutions, we use additional constraints. It can be seen that
\begin{align}
\begin{split}
 [\M n]_\times \M H_j^\top \M H_j [\M n]_\times^\top &= [\M n]_\times  \M R_j^\top \M R_j [\M n]_\times^\top, \\
&= [\M n]_\times [\M n]_\times^\top .\\
 \end{split}
\label{eq:12}
\end{align}
Substituting~\eqref{eq:07} into~\eqref{eq:12} gives
\begin{align}
 [\M n]_\times \M Q_j [\M n]_\times^\top \sim [\M n]_\times [\M n]_\times^\top,
\label{eq:13}
\end{align}
with
\begin{align}
\M Q_j = (\M K_j^{-1} \M G_j \M K_1)^\top (\M K_j^{-1} \M G_j \M K_1).
\label{eq:14}
\end{align}
Note that, we use $\sim$ instead of equality in~\eqref{eq:13} since $\M H_j$ from~\eqref{eq:07} is up to a scale factor. Since $\M Q_j$ are symmetric matrices, we can write them as
\begin{equation}
\M Q_j=\begin{bmatrix}
q_{j1} & q_{j2} & q_{j3}\\
q_{j2} & q_{j4} & q_{j5}\\
q_{j3} & q_{j5} & q_{j6}
\end{bmatrix}.\label{eq:15}
\end{equation}
Then~\eqref{eq:13} can be rewritten as
\begin{align}
\begin{split}
 [\M n]_\times \M Q_j [\M n]_\times^\top =s_j [\M n]_\times [\M n]_\times^\top,\ j=2,3 \\
 \end{split}
\label{eq:16}
\end{align}
where we add scale factors to ensure the equality. Note that both the left and right of~\eqref{eq:16} are symmetric matrices, hence we can get 6 equations for each $j$. 

To simplify our 12 equations, we can eliminate 
some unknowns from these equations using the elimination ideal technique~\cite{cox2005using}. This technique was recently used to solve several minimal camera geometry problems~\cite{kukelova2017clever}.


In our case, we first create an ideal $J$ generated by 12 polynomials~\eqref{eq:16}. 
Then, the unknown elements of the normal $\{n_x,n_y\}$ and the scale factor $s_2,s_3$ are eliminated from the generators of $J$ by computing the generators of the elimination ideal $J_1 = J \cap \mathbb{C}[q_{21}, \dots , q_{36}]$. Here, $q_{j.}$ are the entries of $\M Q_2, \M Q_3$.
These generators can be computed using computer algebra software Macaulay2~\cite{M2} (for more details and the input Macaulay2 code see the SM).



\noindent In this case, the elimination ideal $J$ is generated by seven polynomials $g_i$ of degree 6 in the elements of $\M Q_j$, $j =2,3$. The final constraints are only related to the 12 elements of the symmetric matrices $\M Q_i$ (6 from $\M Q_2$ and 6 from $\M Q_3$).  To the best of our knowledge, these constraints are first shown in this paper and have not been used in the computer vision literature before. Alternatively, we can eliminate $n_x,n_y$ from the 18 equations of the trace constraints~\eqref{eq:11}, however, this will result in more complicated equations since the constraints will be related to the 18 elements of $\M H_i$. 

%% file: sec/5_solvers.tex
\section{New Solvers}

In this section, we propose solvers for the four different cases outlined in Tab.~\ref{tab:cases} using the constraints derived in the previous section. We also propose a method based on LO-RANSAC~\cite{chum2003loransac} that utilizes the new solvers for robust estimation of focal lengths from three views of planar scenes.

\label{sec:solvers}
\subsection{One Unknown Focal Length Parameter}
\label{sec:solvers_one_unknown}
\noindent\textbf{Case \MakeUppercase{\romannumeral1}.}
We first consider the case where the three cameras have equal and unknown focal length, \ie, $f_1=f_2=f_3=f$, and $\M K_{1,2,3} = {\rm diag}(f,f,1)$. By substituting $\M K_j$ into the generators $g_i$, $i=1,\dots,7$ of the elimination ideal $J$, we obtain 7 univariate polynomials in $f$, which are of degree 9 in $\alpha=f^2$. They form an over-constrained system. We only need one of them to find the solutions to $f$. To find the roots of the degree 9 univariate polynomial we use the Sturm sequence method~\cite{gellert2012vnr}. We denote this solver as \h{fff}.

\noindent\textbf{Case \MakeUppercase{\romannumeral2}.}
In the second case, we assume that $f_1$ is known, and $f_2=f_3=f$ are unknown. This case occurs when a calibrated camera is used to capture the reference image, and the second uncalibrated camera captures two target images. Similar to \case{1}, we only need to find the roots of a univariate polynomial in $f$, in this case of degree 6, using Sturm sequences~\cite{gellert2012vnr}. We denote this solver as \h{ff}.

\subsection{Two Unknown Focal Length Parameters}
\label{sec:solvers_two_unknown}
In the second group of solvers, we consider self-calibration problems with two different unknown focal lengths. There are two practical cases:

\noindent\textbf{Case \MakeUppercase{\romannumeral3}.}
In this case, we assume two unknown focal lengths
$f_1=f$, and $f_2=f_3=\rho$, \ie, $\M K_1 = {\rm diag}(f,f,1)$ and $\M K_{2,3} = {\rm diag}(\rho,\rho,1)$. This corresponds to a situation where the first uncalibrated camera is used to capture the reference image and the second uncalibrated camera captures two target images. By substituting $\M K_j$ into the constraints $g_i$ from the elimination ideal, we  obtain 7 polynomials in two unknowns 
$\alpha,\beta,(\alpha=f^2,\beta = \rho^2)$, 
which can be written as
\begin{equation}
{\M M}{\M v} =\M 0, \label{eq:17}
\end{equation}
where ${\M M}$ is a $7\times 28$ coefficient matrix and
\begin{equation}
{\M v}=[1,\beta,...,\beta^6,\alpha,\alpha\beta,...,\alpha\beta^6,...,\alpha^3,...,\alpha^3\beta^6]^\top,\label{eq:18}
\end{equation}
is a vector consisting of the 28 monomials. The system of polynomial equations in~\eqref{eq:17} can be solved using different algebraic methods~\cite{cox2005using}. There are also different state-of-the-art approaches for generating efficient algebraic solvers~\cite{kukelova2008automatic,larsson2018beyond, Bhayani_2020_CVPR, kukelova2012polynomial, hartley2012efficient}. In this paper, we use the hidden variable technique to derive polynomial eigenvalue solution based on~\cite{kukelova2012polynomial}.

\noindent\textbf{Polynomial Eigenvalue Solution.}\label{sec:pep}
The polynomial system in~\eqref{eq:17} contains four polynomials in two unknowns $(\alpha,\beta)$, and the highest degree of the unknown $\alpha$ is 3. In this case, $\alpha$ can be chosen as the hidden variable, \ie we can consider it as a parameter. Then the system of polynomial equations~\eqref{eq:17}  can be rewritten as 
\begin{equation}
{\M C}(\alpha){\tilde {\M v}} =\M 0, \label{eq:19}
\end{equation}
where ${\M C}(\alpha)$ is a $7\times7$ polynomial matrix parameterized by $\alpha$, and ${\tilde {\M v}}=[1,\beta,...,\beta^6]^{\top}$ is a vector of monomials in $\beta$ without $\alpha$. ${\M C}(\alpha)$ can be rewritten as
\begin{equation}
{\M C}(\alpha) = \alpha^{3}{\M C}_{3} + \alpha^{2}{\M C}_{2} + \alpha{\M C}_1 + {\M C}_0, \label{eq:20}
\end{equation}
where ${\M C}_{3},{\M C}_{2},{\M C}_1,{\M C}_0$ are  $7\times7$ coefficient matrices containing only numbers. For this problem, the matrix $\M C_3$ is only rank 4, resulting in four zero eigenvalues. 
 To speed up the solver, we remove these zero eigenvalues from the computations. To do this, we first need to transform the matrices $\M C_i$ by considering linear combinations of their rows, such that there are three zero rows in the transformed matrix $\M C_3$.
To remove the zero rows in the transformed $\M C_3$, 
we use the technique from~\cite{ding2020homography} and consider the transpose of~\eqref{eq:19}
\begin{equation}
{\tilde {\M v}^\top}{\M C^\top}(\alpha) =\M 0. \label{eq:21}
\end{equation}
In this case, we have
\begin{equation}
{\M C^\top}(\alpha) = \alpha^{3}{\M C}_3^\top + \alpha^{2}{\M C}_2^\top + \alpha{\M C}_1^\top + {\M C}_0^\top. \label{eq:22}
\end{equation}
The zero rows in ${\M C}_3$ are now zero columns in ${\M C}_3^\top$. Since ${\M C}_0$ is full rank, we let $\gamma = \frac{1}{\alpha}$ and rewrite~\eqref{eq:22} as
\begin{equation}
{\M C^\top}(\gamma) = \gamma^{3}{\M C}_0^\top + \gamma^{2}{\M C}_1^\top + \gamma{\M C}_2^\top + {\M C}_3^\top. \label{eq:23}
\end{equation}
If we consider~\eqref{eq:23} as a polynomial eigenvalue problem~\cite{bai2000templates}, the solutions to $\gamma$ are the eigenvalues of $21\times 21$ matrix
\begin{eqnarray}
\M D = \begin{bmatrix}
\M 0 & {\M I} & \M 0 \\
\M 0 & \M 0 & {\M I}  \\
 -{\M C}_0^{-\top}{\M C}_3^\top & -{\M C}_0^{-\top}{\M C}_2^\top & -{\M C}_0^{-\top}{\M C}_1^\top
\end{bmatrix}.\label{eq:24}
\end{eqnarray}
The three zero columns in ${\M C}_3^\top$ 
can now be removed together with their corresponding rows to eliminate the zero eigenvalues~\cite{kukelova2012polynomial}. In this way, we obtain 18 possible solutions. Once we have solutions to $\alpha$, the remaining unknown $\beta$ can be extracted from the null vector of $\M C(\alpha)$ based on~\eqref{eq:19}. We denote this solver as \h{f \rho \rho}.

Note that in the proposed polynomial eigenvalue formulation, we solve a relaxed version of the original problem~\eqref{eq:11}. The original system~\eqref{eq:11} has 17 solutions, as it can be shown, \eg, using the computer algebra system Macaulay2~\cite{M2}. In the polynomial eigenvalue formulation, we have one spurious solution that does not ensure that the elements of $\tilde {\M v}$ satisfy ${\tilde {\M v}}=[1,\beta,\cdots,\beta^6]^{\top}$.




\noindent\textbf{Case \MakeUppercase{\romannumeral4}.} Finally, we consider a case where $f_1$ is known and $f_2=f$, $f_3=\rho$ are unknown. The solver for this case, we denote it as \h{f \rho}, performs steps similar to the solver \h{f \rho \rho} for \case{3}. In this case, the \h{f \rho} solver computes the eigenvalues of a $12 \times 12$ matrix. Due to space limitations, we describe the \h{f \rho} solver for \case{4} in the SM.

\subsection{Robust Estimation of Focal Lengths}

\label{sec:robust_estimation}



To estimate focal lengths from three images of planar scenes we utilize the LO-RANSAC framework~\cite{chum2003loransac} using a strategy inspired by~\cite{nister2004efficient}. 
We first extract triplet point correspondences from images (e.g. using~\cite{detone2018superpoint, lindenberger2023lightglue}). In each RANSAC iteration we sample 4 triplets from which we estimate $\M G_2$ and $\M G_3$ using DLT~\cite{hartley2003multiple}. 
These matrices are used as inputs to solvers proposed in Section~\ref{sec:solvers_one_unknown} and~\ref{sec:solvers_two_unknown}. 

For each resulting real positive solution we use \eqref{eq:07} to obtain $\M H_2$, which is then decomposed into two possible poses $(\M R_2, \M t_2)$. 
We then use these poses and focal lengths to triangulate three of the sampled points in the first two views thus obtaining points $\M X_{i1}$. 
We use the corresponding points in the third view to obtain $\M R_3$ and $\M t_3$ using the \pnp{3} solver~\cite{ding2023revisiting}. 
Note that it is possible to obtain $\M R_3$ and $\M t_3$ by decomposing $\M H_3$ and using a single correspondence to obtain the scale of $\M t_3$, but in practical experiments we found the approach using \pnp{3} faster and more accurate. We score the generated models using pairwise Sampson error for each of the three image pairs. Whenever a new so-far-best model is found we perform local optimization using the Levenberg-Marquardt algorithm minimizing the pairwise Sampson error across the three pairs of views. In addition to the estimated focal lengths this strategy also produces the relative poses of all three cameras.
We denote the proposed robust estimators as \hr{fff} for \case{1}, \hr{ff} for \case{2}, \hr{f \rho \rho} for \case{3} and \hr{f \rho} for \case{4} to distinguish them from the solvers.
Our implementation is based on PoseLib~\cite{poselib}. 

For \case{1} and II this approach is able to find the single unknown focal length even when the scene is fully planar. For \case{3} and IV we may obtain multiple solutions from which the correct focal lengths can not be distinguished using only planar points thus requiring some off-plane points during scoring and local optimization.

%% file: sec/6_synthetic.tex
\section{Experiments}

We perform extensive experiments on synthetic and real data to evaluate the performance of the proposed solvers and the robust focal length estimation strategy described in Section~\ref{sec:robust_estimation}. 
%
We compare our solvers with several baselines using either pairwise or triplet correspondences. For pairwise correspondences, we consider the 6 point solver for relative pose with one unknown shared focal length~\cite{kukelova2012polynomial} denoted as \fEf and its combination with the $4 + 1$ point plane and parallax solver~\cite{torii2011six} using the DEGENSAC framework~\cite{chum2005degensac}, which we denote as \fEfp. We also evaluate the solver for one unknown focal length~\cite{bujnak20093d} denoted as \Ef. When considering triplets, we use a strategy~\cite{nister2004efficient} of using a pairwise solver 
followed by triangulation of points and registration of the third camera using the $\M P3\M P$ solver~\cite{ding2023revisiting}. 
(see Section~\ref{sec:robust_estimation} for more details). We denote the methods that work with triplet correspondences as \fEfr, \fEfpr, \Efr\xspace respectively.

Note that we do not compare with the solvers proposed in~\cite{heikkila2017using}. As visible from Table~\ref{tab:cases} these solvers are significantly slower than our solvers~\footnote{The reported runtimes are runtimes of the original Matlab implementations of~\cite{heikkila2017using} and the Matlab + mex implementations of our solvers on I7-11700K CPU.} and they return significantly more solutions that need to be tested inside RANSAC. As such they are not practical for real-world applications. Moreover, the solver for \case{4} proposed in~\cite{heikkila2017using} returns solutions only to one of the two unknown focal lengths.

Except for the numerical stability experiment, due to space constraints, in the main paper we present only results for  \case{1} and  \case{2}. Experiments for \case{3} and IV for synthetic and real data are in the SM.

\subsection{Synthetic experiments}

\begin{figure}[t]
\begin{center}
\subfloat{
\includegraphics[width = 0.5\linewidth]{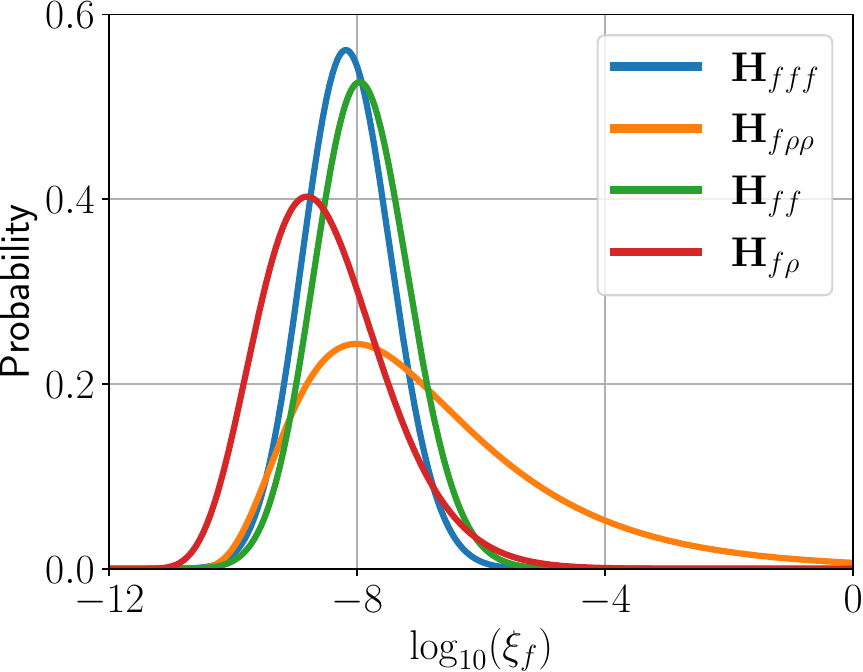}}
\end{center}
\vspace{-2em}
\caption{Numerical stability of the proposed solvers.}\label{fig:stability}
\end{figure}

\begin{figure*}
    \centering
\resizebox{\textwidth}{!}{
\begin{tikzpicture} 
    \begin{axis}[%
    hide axis, xmin=0,xmax=0,ymin=0,ymax=0,
    legend style={draw=white!15!white, 
    line width = 1pt,
    legend  columns =-1, 
    /tikz/every even column/.append style={column sep=0.2cm}
    }
    ]

    \addlegendimage{line width=4pt, Seaborn1}
    \addlegendentry{\hr{fff} (ours)};
    \addlegendimage{line width=4pt, Seaborn2}
    \addlegendentry{\hr{ff} (ours)};
    \addlegendimage{line width=4pt, Seaborn3}
    \addlegendentry{\fEfr~\cite{tzamos2023relative}};
    \addlegendimage{line width=4pt, Seaborn4}
    \addlegendentry{\fEfpr};
    \addlegendimage{line width=4pt, Seaborn5}
    \addlegendentry{\Efr}; 
    \addlegendimage{line width=4pt, Seaborn6}
    \addlegendentry{\fEf~\cite{kukelova2012polynomial}}; 
    \addlegendimage{line width=4pt, Seaborn7}    \addlegendentry{\fEfp~\cite{torii2011six}}; 
    \addlegendimage{line width=4pt, Seaborn8}
    \addlegendentry{\Ef~\cite{bujnak20093d}};

    \end{axis}
    \end{tikzpicture}}

    \begin{subfigure}{0.30\textwidth}
    \includegraphics[width=\textwidth]{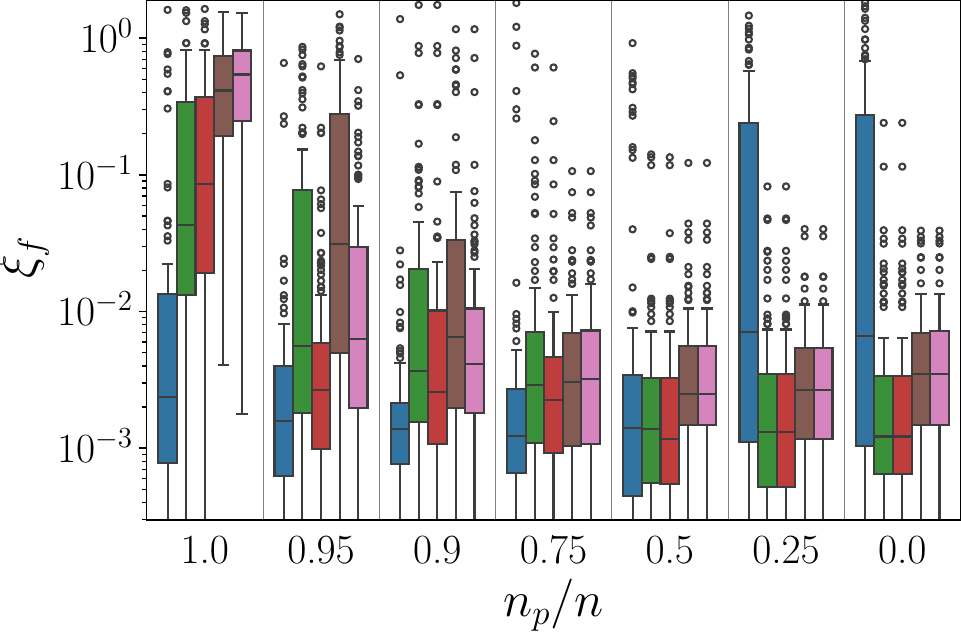}
    \caption{}
    \end{subfigure}
    \hfill    
    \begin{subfigure}{0.30\textwidth}
    \includegraphics[width=\textwidth]{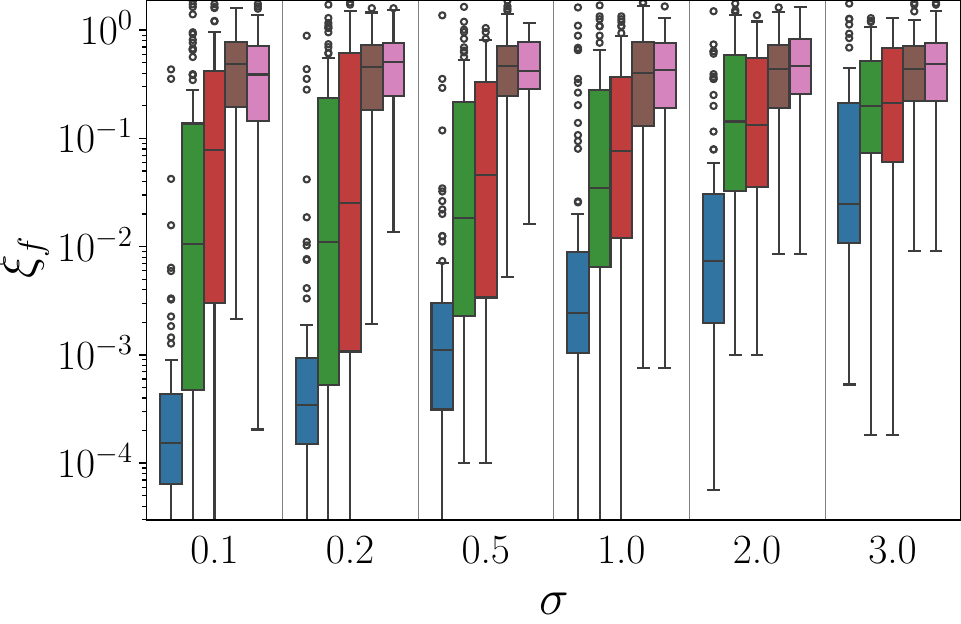}
    \caption{}
    \end{subfigure}
    \hfill    
    \begin{subfigure}{0.30\textwidth}
    \includegraphics[width=\textwidth]{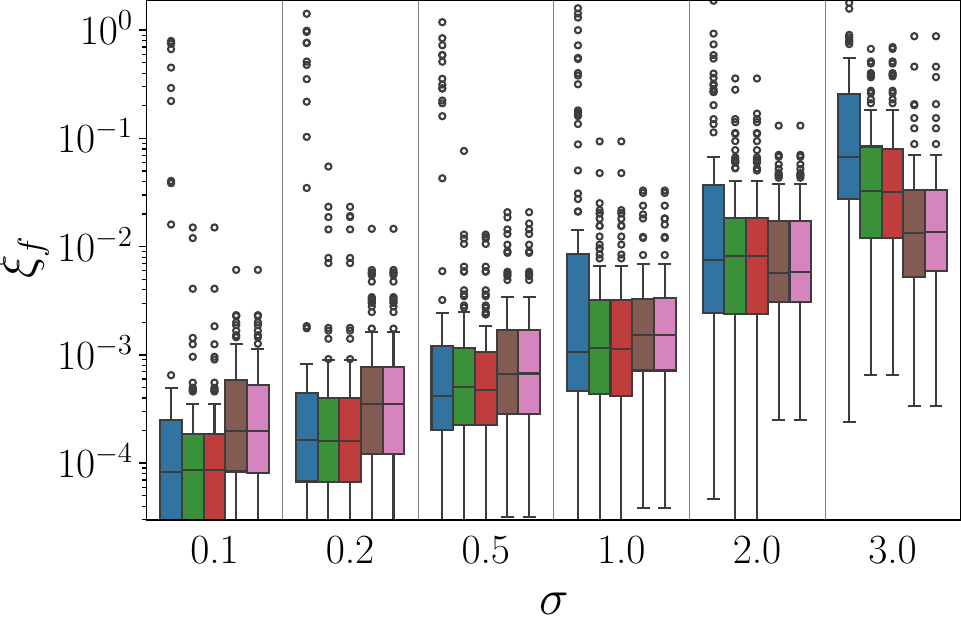}
    \caption{}
    \end{subfigure}

    \vspace{0.3em}
   
    \begin{subfigure}{0.30\textwidth}
    \includegraphics[width=\textwidth]{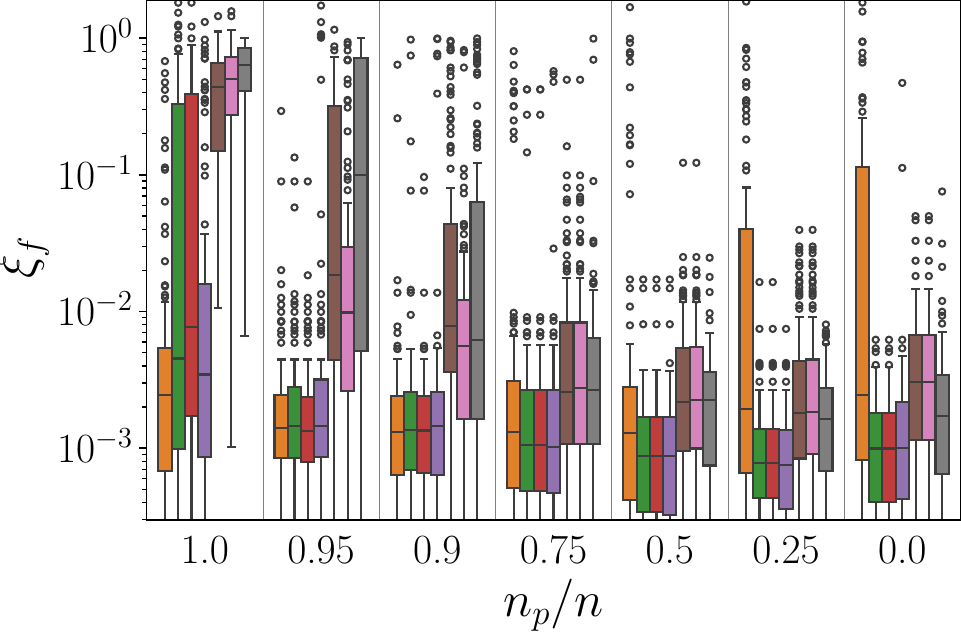}
    \caption{}
    \end{subfigure}
    \hfill    
    \begin{subfigure}{0.30\textwidth}
    \includegraphics[width=\textwidth]{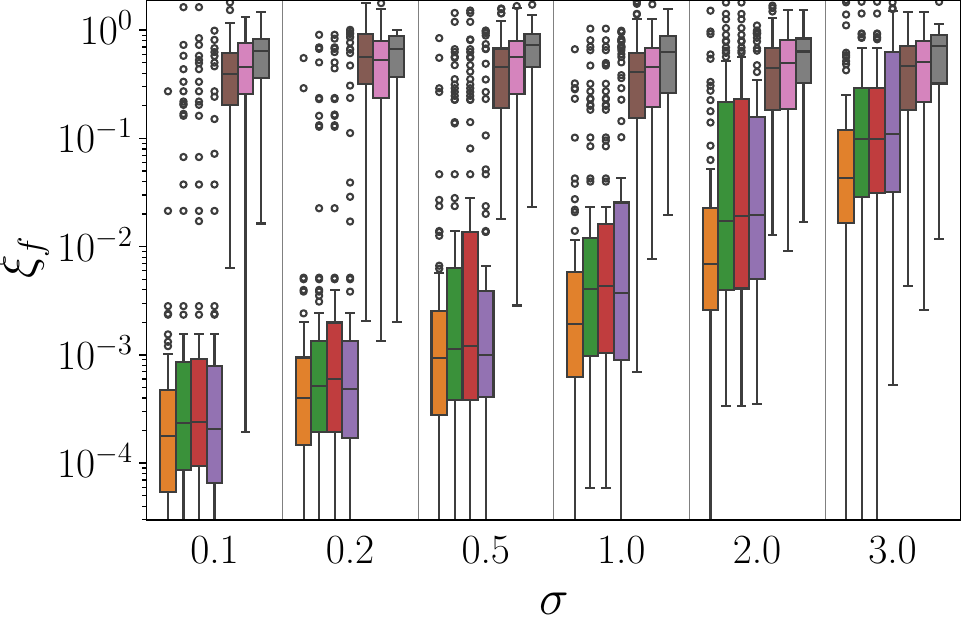}
    \caption{}
    \end{subfigure}
    \hfill    
    \begin{subfigure}{0.30\textwidth}
    \includegraphics[width=\textwidth]{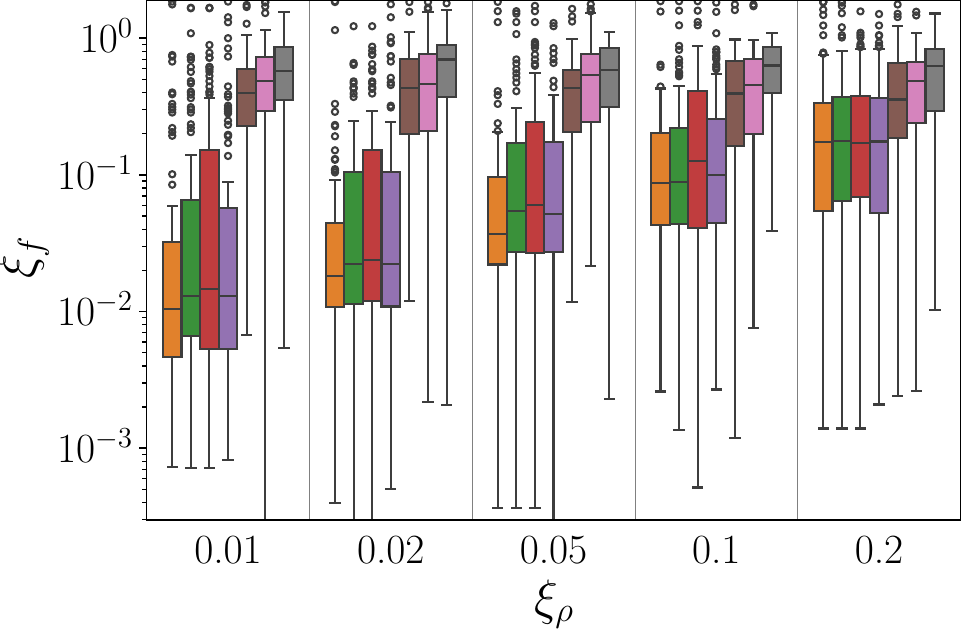}
    \caption{}
    \end{subfigure}

    \caption{Focal length errors for the evaluated methods in synthetic experiments. \textbf{\case{1}}: (a) We vary the proportion of points which lie on the dominant plane with fixed noise $\sigma = 1$. (b, c) We vary noise $\sigma$ with (b) $n_p / n = 1.0$ (\ie all points lie on a plane) and (c) $n_p / n = 0.5$. \textbf{\case{2}}: (d) We vary the proportion of points which lie on the dominant plane with fixed noise $\sigma = 1$. (e) We vary noise $\sigma$ with $n_p / n = 1.0$. (f) We perturb the known focal length $\rho$ so that its error is $\xi_\rho$ with fixed noise $\sigma = 1$ and $n_p / n = 1.0$.}
    \label{fig:synth_accuracy}
\end{figure*}

\noindent\textbf{Numerical Stability.}
In the first experiment, we study the numerical stability of the solvers proposed in Sections \ref{sec:solvers_one_unknown} and \ref{sec:solvers_two_unknown}. The synthetic data is generated in the following setup. We sample 200 3D points distributed on a plane with a random orientation. The focal lengths of the cameras are generated from a uniform distribution $f_g \in [300,3000]$ px with a field of view of 90 degrees. The baseline between consecutive cameras is set to be 10 percent of the average scene distance. We generated 10,000 random scenes with 3D points on different planes and different transformations between consecutive views. The focal length error $\xi_f$ is defined as:
\begin{equation}
\xi_f=\frac{|f_e-f_g|}{f_g},
\label{eqn:f_err}
\end{equation}
where $f_g, f_e$ represent the ground-truth focal length and the estimated focal length, respectively. For the solvers with different focal lengths, we use the geometric mean of the two focal length errors $\xi_f=\sqrt{\xi_{f_1} \xi_{f_2}}$. 

Fig.~\ref{fig:stability} shows the results of the ${\rm log}_{10}$ relative focal length error for the proposed methods by considering the solution closest to the ground truth. As can be seen, all of our solvers are numerically stable without large errors. 

\PAR{Accuracy of the Estimated Focal Lengths.}
Next, we evaluate the performance of the robust estimators proposed in Section~\ref{sec:robust_estimation} on synthetic data. We show how varying noise levels and the proportion of points that lie on a plane affect their accuracy compared to baselines.


\begin{table*}[t]
\centering
\resizebox{0.68\linewidth}{!}{
\begin{tabular}{|c|rc|c|ccccc|} \cline{2-9}
\multicolumn{1}{c|}{} & \multicolumn{2}{|c|}{Method} & Sample & Median $\xi_f$ & Mean $\xi_f$ & mAA$_f$(0.1) & mAA$_f$(0.2) & Runtime (ms) \\ \hline
\multirow{ 5 }{*}{\rotatebox[origin=c]{90}{\case{1}}}
 & \hr{fff} & \textbf{ours} & 4 triplets & \textbf{0.0439}&\textbf{0.1491}&\textbf{51.49}&\textbf{65.58}&49.31 \\ 
 & \fEfr & \cite{tzamos2023relative} & 6 triplets & 0.0503&0.2215&47.83&61.45&37.44 \\ 
 & \fEfpr &  & 6 triplets & 0.0518&0.2240&47.11&60.65&41.18 \\ 
 & \fEf & \cite{kukelova2012polynomial} & 6 pairs & 0.4793&0.7143&10.49&17.07&\textbf{24.54} \\ 
 & \fEfp & \cite{torii2011six} & 6 pairs & 0.5166&0.7935&10.31&16.51&31.95 \\ 
\hline
\multirow{ 7 }{*}{\rotatebox[origin=c]{90}{\case{2}}}
 & \hr{ff} & \textbf{ours} & 4 triplets & \textbf{0.0611}&\textbf{0.2244}&\textbf{43.11}&\textbf{55.54}&35.97 \\ 
 & \fEfr & \cite{tzamos2023relative} & 6 triplets & 0.0692&0.2438&40.42&52.35&31.81 \\ 
 & \fEfpr &  & 6 triplets & 0.0714&0.2529&39.86&51.58&34.35 \\ 
 & \Efr &  & 6 triplets & 0.0691&0.2655&40.60&52.13&36.82 \\ 
 & \fEf & \cite{kukelova2012polynomial} & 6 pairs & 0.4623&0.7085&12.01&19.12&28.59 \\ 
 & \fEfp & \cite{torii2011six} & 6 pairs & 0.4968&0.7875&12.11&18.98&37.60 \\ 
 & \Ef & \cite{bujnak20093d} & 6 pairs & 0.6827&0.6843&\phantom{1}8.87&13.01&\textbf{22.59} \\ 
\hline
\end{tabular}
}
\caption{Comparison of the focal length estimation accuracy of the evaluated methods on our dataset of planar scenes.}
\label{tab:eval}
\end{table*}

\begin{figure*}
    \centering
        \resizebox{0.7\linewidth}{!}{
        \begin{tikzpicture} 
        \begin{axis}[%
        hide axis, xmin=0,xmax=0,ymin=0,ymax=0,
        legend style={draw=white!15!white, 
        line width = 1pt,
        legend  columns =5, 
        /tikz/every even column/.append style={column sep=0.5cm},
        }
        ]

        \addlegendimage{Seaborn1}
        \addlegendentry{\hr{fff} (ours)};
        \addlegendimage{Seaborn2}
        \addlegendentry{\hr{ff} (ours)};
        \addlegendimage{Seaborn3}
        \addlegendentry{\fEfr};
        \addlegendimage{Seaborn4}
        \addlegendentry{\fEfpr};
        \addlegendimage{Seaborn5}
        \addlegendentry{\Efr};    
        
        \end{axis}
    \end{tikzpicture}}

    \vspace{-0.2em}

    \begin{subfigure}{0.49\textwidth}
    \includegraphics[width=0.49\linewidth]{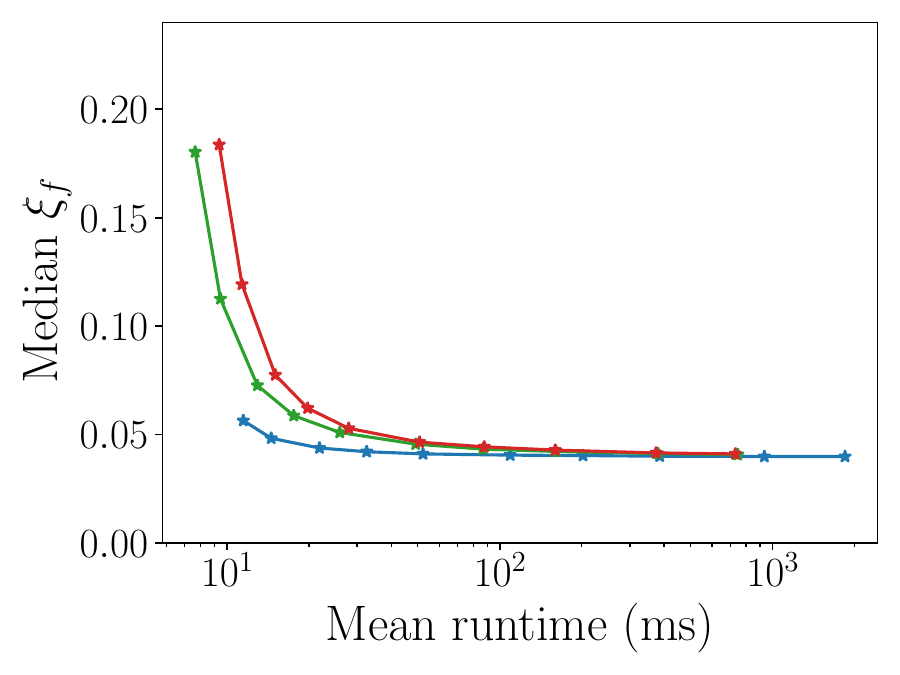}  \hfill \includegraphics[width=0.49\linewidth]{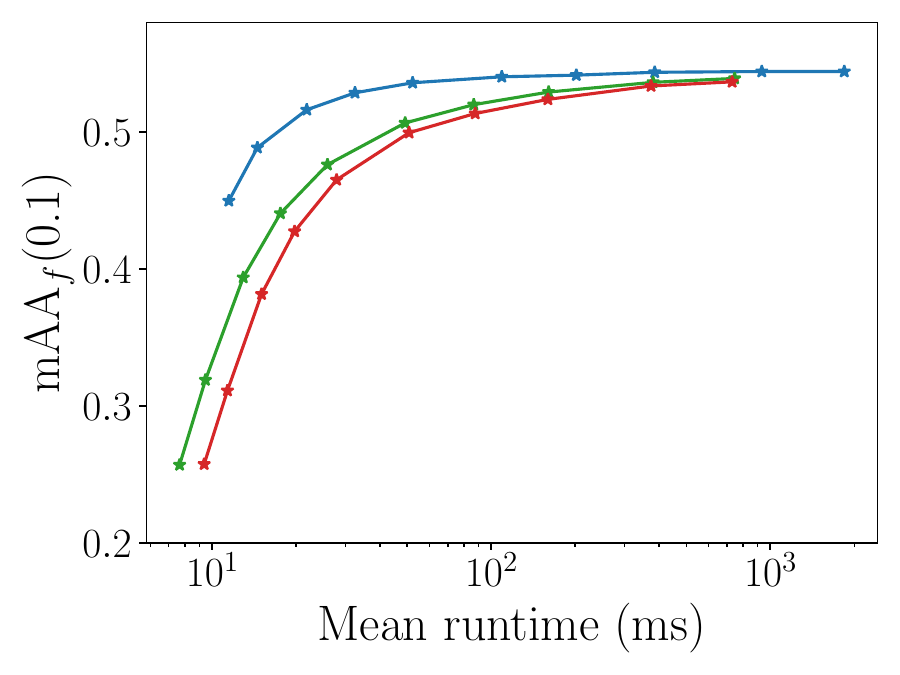}  
    \vspace{-0.5em}
    \caption{\case{1}}    
    \label{fig:eval_graph_case_1}
    \end{subfigure}
    \hfill    
    \begin{subfigure}{0.49\textwidth}
    \includegraphics[width=0.49\linewidth]{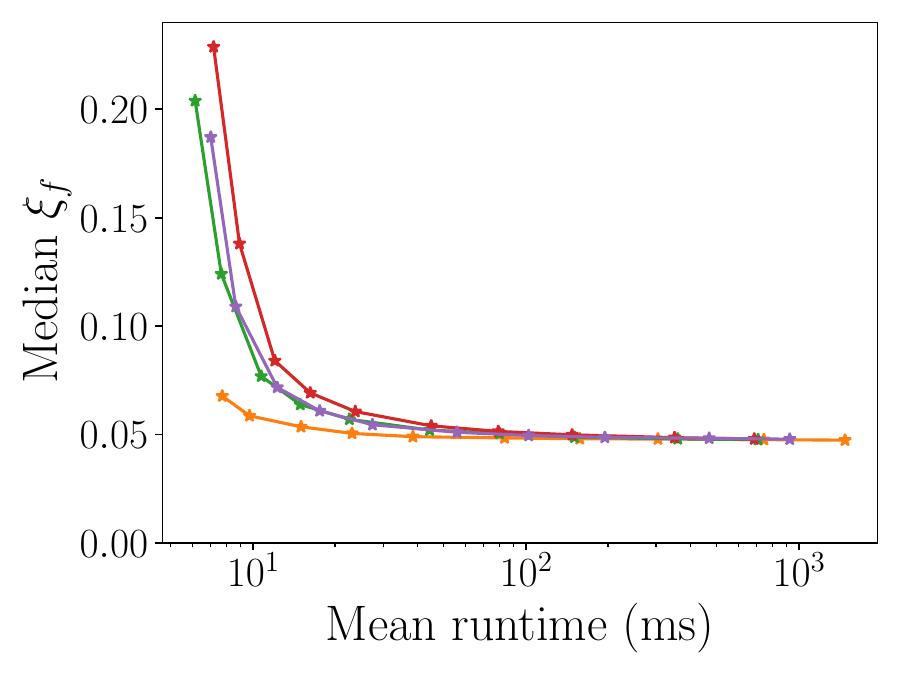}  \hfill \includegraphics[width=0.49\linewidth]{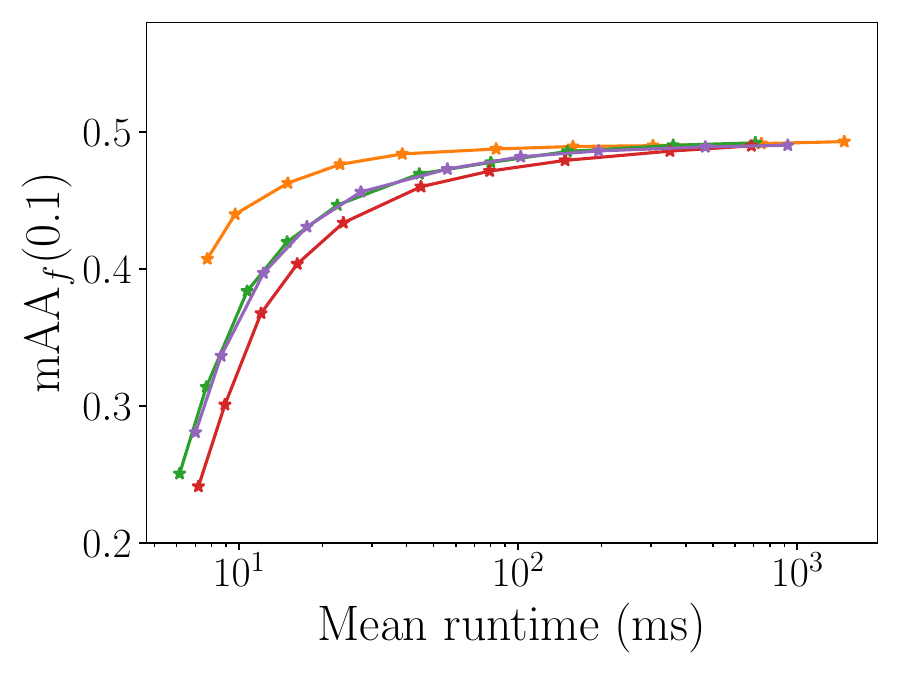}  
    \vspace{-0.5em}
    \caption{\case{2}}    
    \label{fig:eval_graph_case_2}
    \end{subfigure}
    \caption{Median $\xi_f$ and mAA$_f(0.1)$ plotted for different number of RANSAC iterations (10, 20, 50, 100, 200, 500, 1000, 2000, 5000, 10000). All methods are evaluated in combination with \pnp{3}~\cite{ding2023revisiting} and non-linear optimization within PoseLib~\cite{poselib}. We do not include methods relying solely on pairwise correspondences since they all result in very low accuracy (median $\xi_f>0.4,  \text{mAA}_f(0.1)<0.2$).}
    \label{fig:eval_speed_accuracy}
\end{figure*}

The synthetic data is generated with $n = 200$ 3D points visible by the three $1920 \text{px}\times1080\text{px}$ cameras with focal lengths sampled unfiormly $f_g \in \left[ 300, 3000\right]$. The cameras are positioned randomly such that their views overlap. $n_p$ of the total $n$ points lie on a plane with random orientation. From the 3D points we create triplets with 75\% inlier ratio by randomly shuffling 25\% of the points. We add Gaussian noise with standard deviation of $\sigma$ to the pixel coordinates of points. We run all of the tested methods in PoseLib~\cite{poselib} for a fixed number of 100 iterations with the Sampson error threshold of 3 px. For methods that use only pairwise correspondences we consider only the first two views. For each configuration we generate 100 random scenes.

Fig.~\ref{fig:synth_accuracy} shows the results of the synthetic experiments in which we vary both $\sigma$ and $n_p$. For \case{2} we also show how the error in the known focal length $\rho$ affects the accuracy of the methods (Fig.~\ref{fig:synth_accuracy} (f)). 
When considering \case{1}, \hr{fff} shows significantly better accuracy than the remaining methods when a large portion of points lie on a single dominant plane (Fig.~\ref{fig:synth_accuracy} (a)) even under high levels of noise (Fig.~\ref{fig:synth_accuracy} (b)). 
This indicates a clear superiority of \hr{fff} when dealing with images of planar scenes. Similarly, for \case{2}, \hr{ff} also shows improved focal length estimation accuracy when considering planar scenes, although the improvement over competing methods is not as significant.
For fully planar scenes, as expected, the two-view solvers \fEf, \fEfp\xspace and \Ef\xspace fail to generate correct solutions (Fig.~\ref{fig:synth_accuracy} (b),(e),(f)).
For scenes with few of the plane points, \eg, scenes with $n_p / n \in \left[0.95,0.9\right]$, the \fEfp\xspace solver, \ie, the solver where the $4+1$ plane+parallax solver is utilized within DEGENSAC with \fEf\xspace~\cite{torii2011six}, outperforms the \fEf\xspace solver (Fig.~\ref{fig:synth_accuracy} (a)). This is not surprising since the \fEfp\xspace solver~\cite{torii2011six} was developed to handle scenes with dominant planes and some off-the plane points.
The solvers that work with triplet correspondences, \ie \fEfr, \fEfpr, \Efr, outperform the two-view solvers.



%% file: sec/7_real.tex
\subsection{Real-World Experiments}



\begin{figure}[ht]
    \centering

    \begin{tabular}{cccccc}
    \centering

    \includegraphics[width=0.13\textwidth]{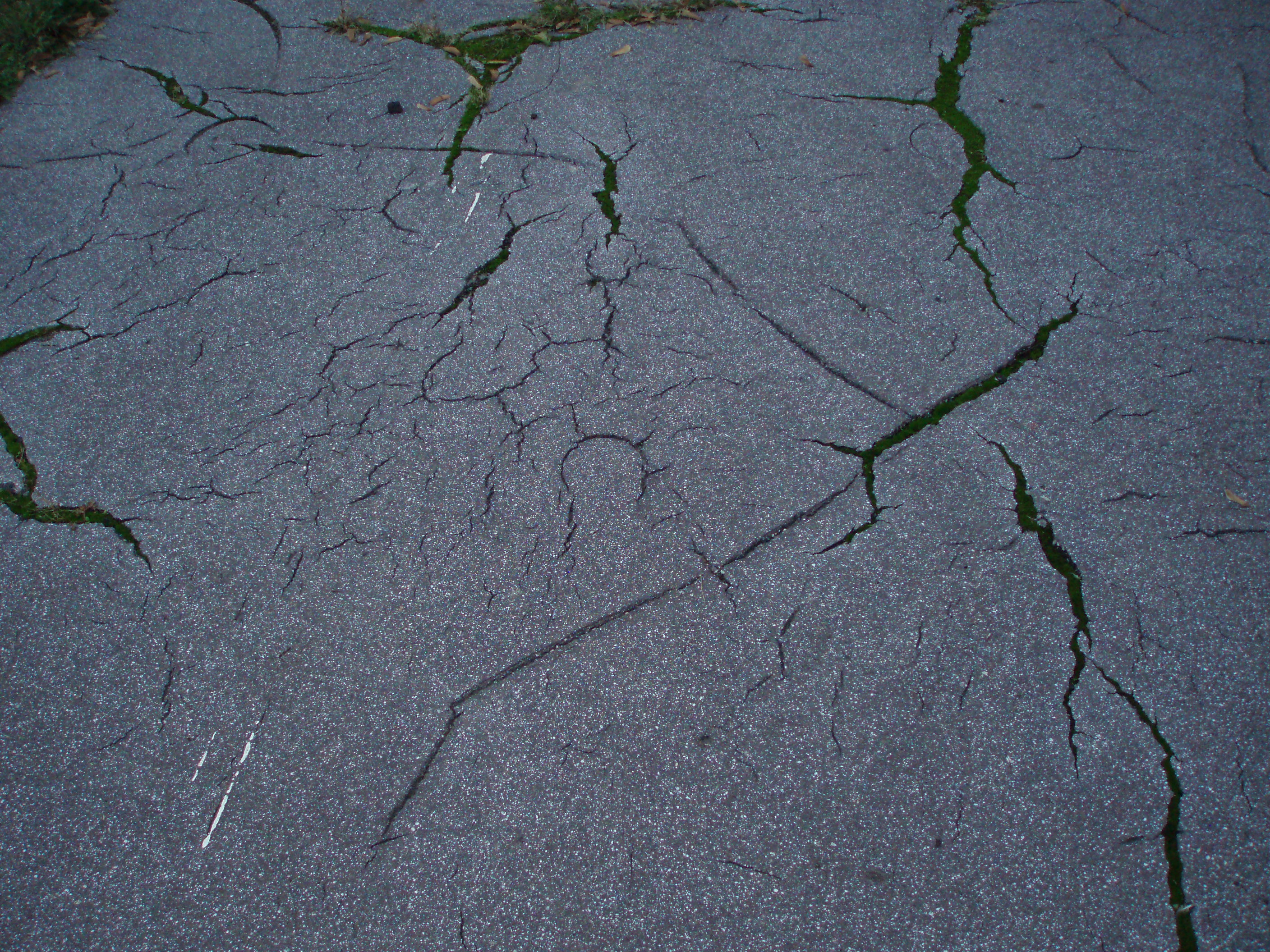} & 
    \includegraphics[width=0.13\textwidth]{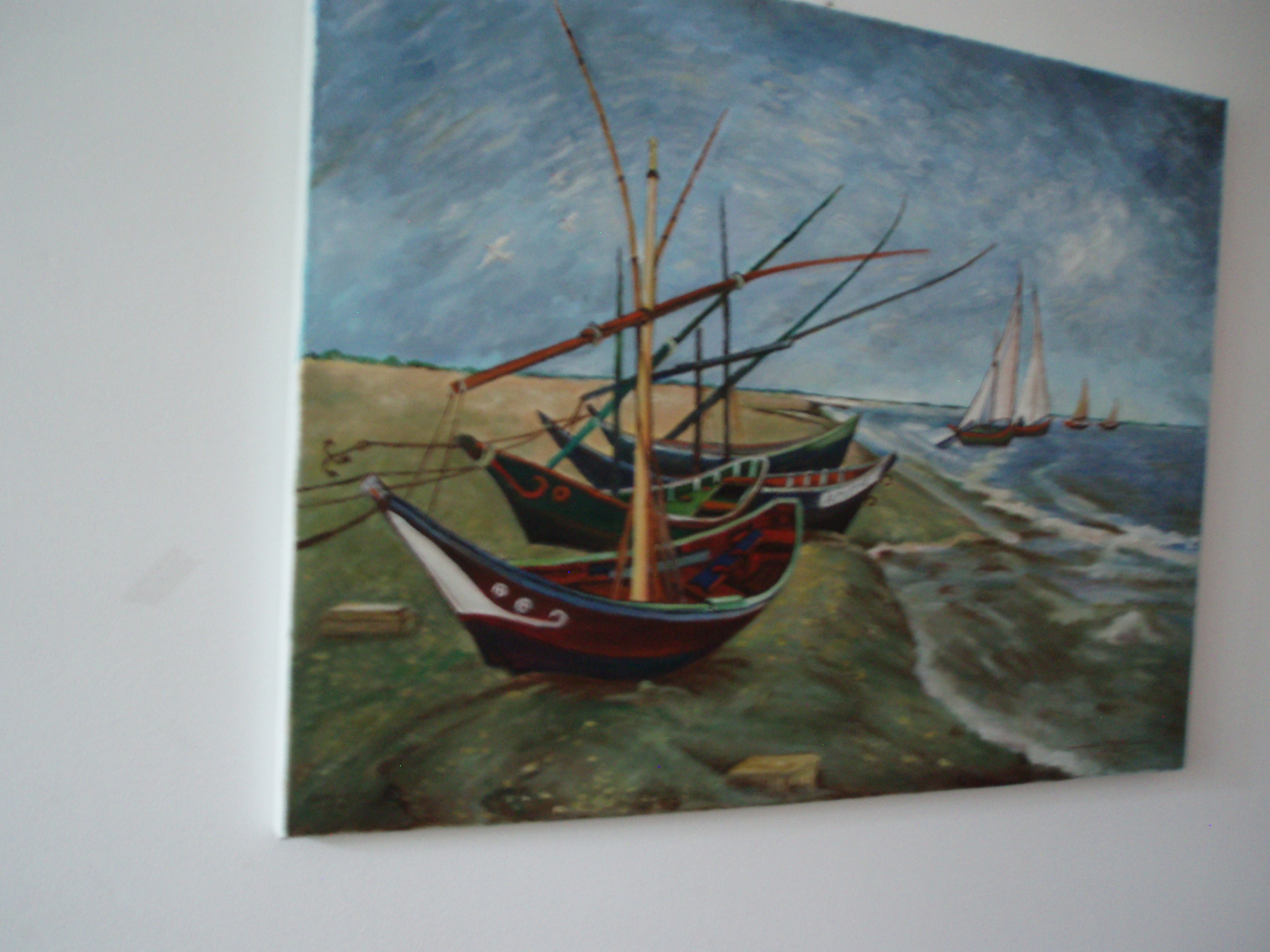} &
    \includegraphics[width=0.13\textwidth]{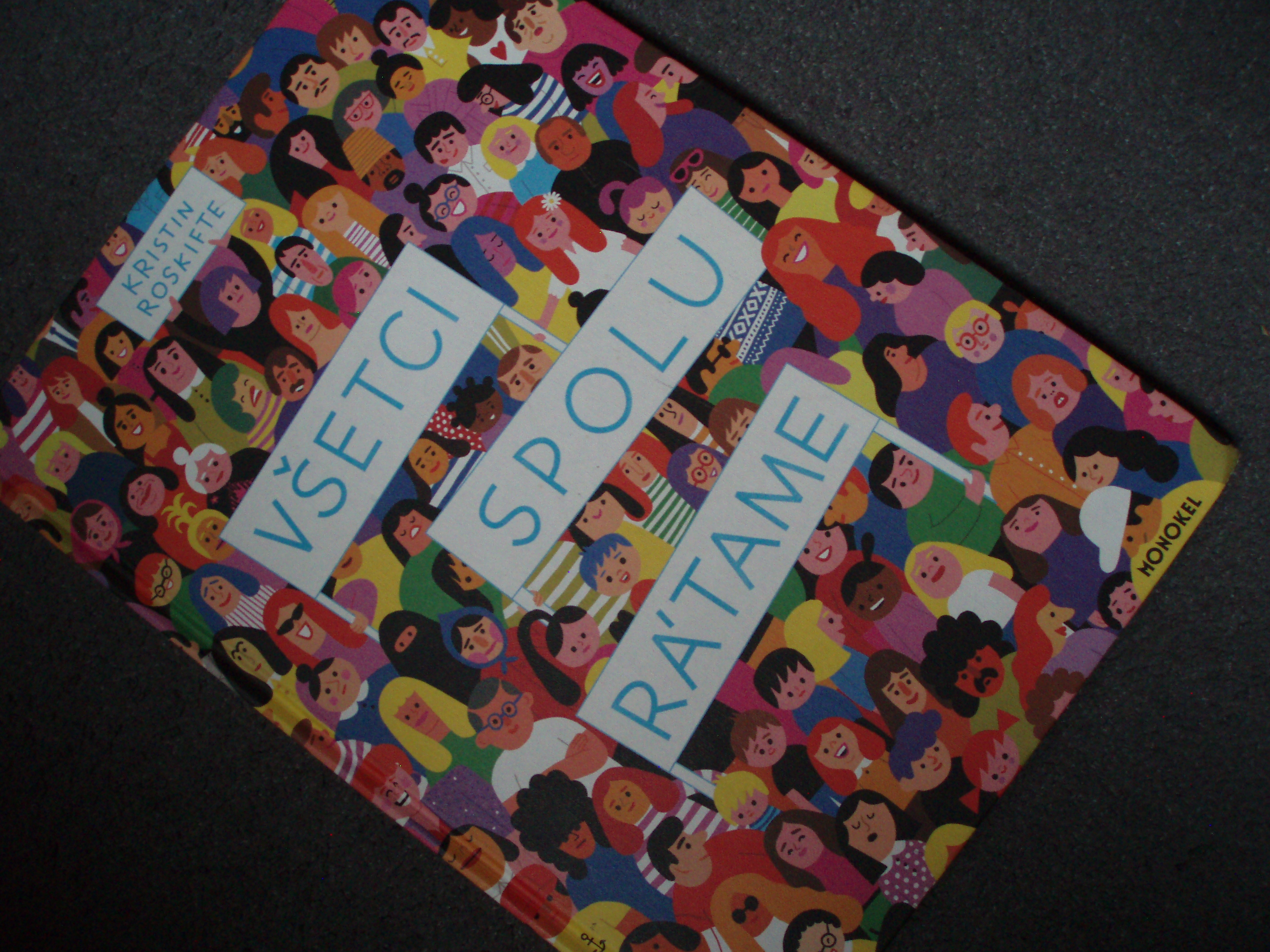} \\
    \small{Asphalt} & \small{Boats} & \small{Book} \\
    \includegraphics[width=0.13\textwidth]{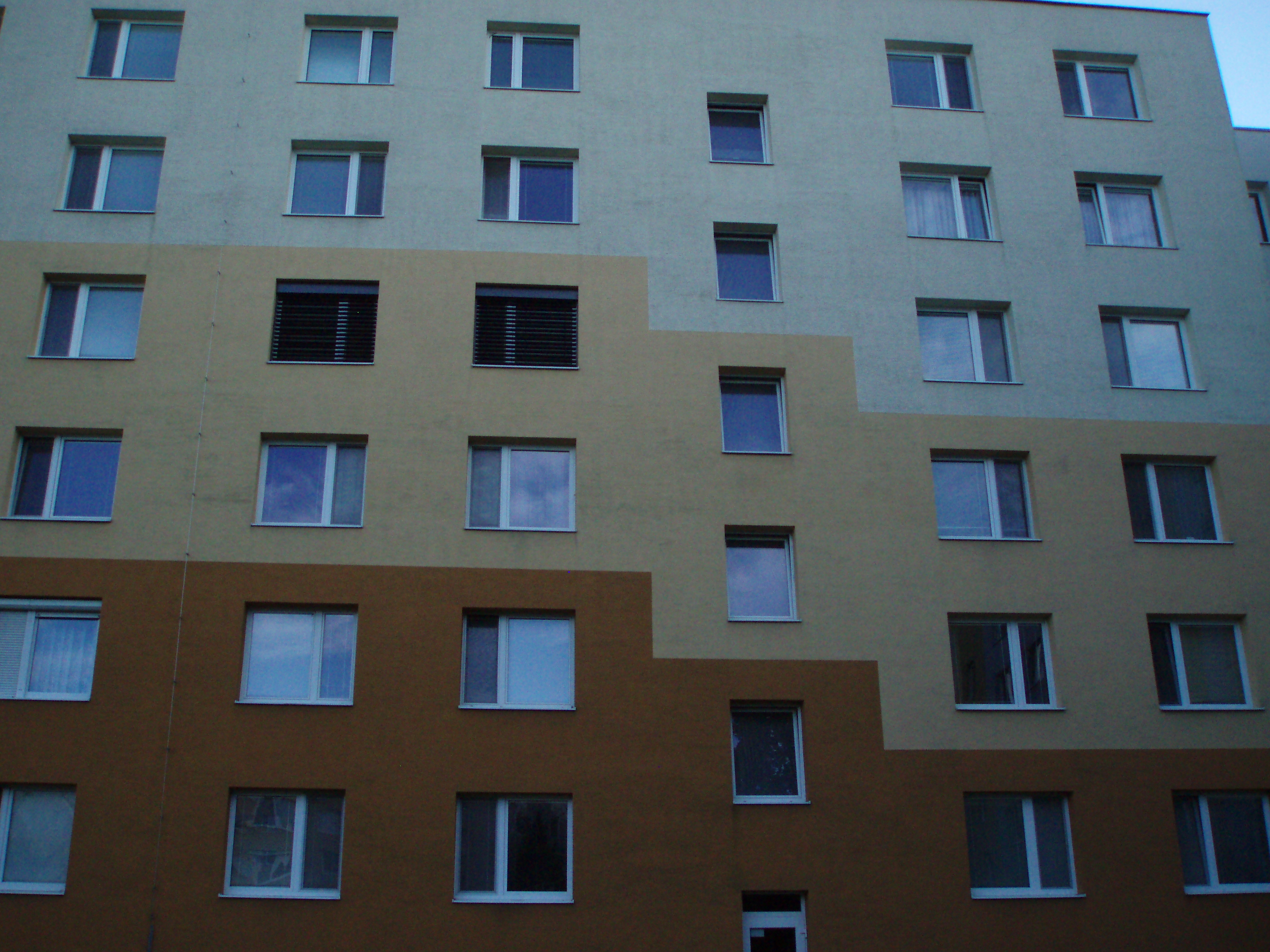} &
    \includegraphics[width=0.13\textwidth]{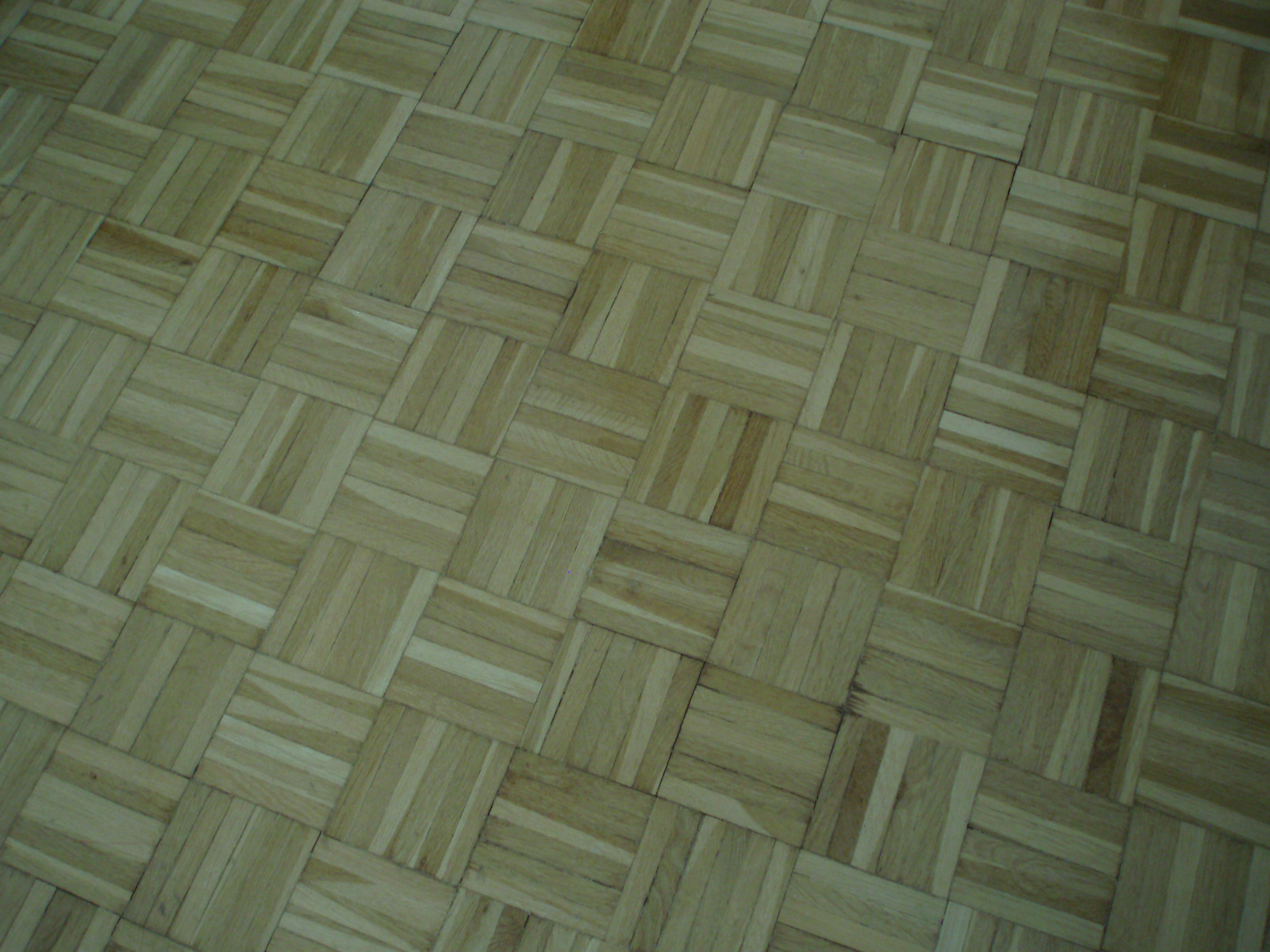} &
    \includegraphics[width=0.13\textwidth]{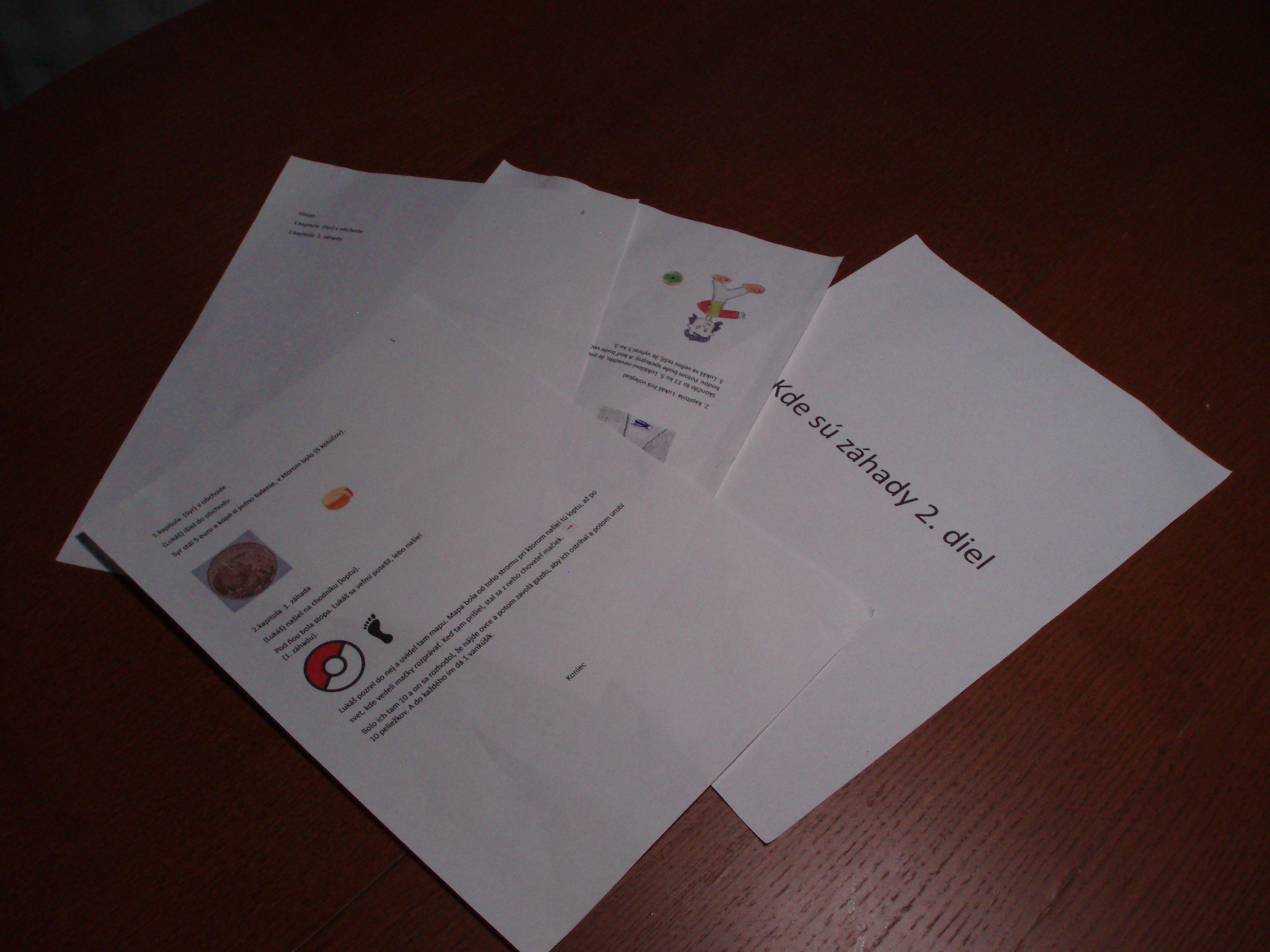} \\
    \small{Facade} & \small{Floor} & \small{Papers}
    \end{tabular}    
    \caption{The six planar scenes captured in our evaluation dataset.}
    \label{fig:dataset_scenes}
\end{figure}

\PAR{Dataset}

Since there are no suitable datasets for evaluating focal length recovery on planar scenes, we have collected a new dataset for evaluation which we will make publicly available. We have collected the dataset using 14 different cameras. For each camera we obtained ground truth focal lengths using the standard calibration method \cite{zhang2000calibration} with a checkerboard pattern. Using these cameras we have captured four indoor and two outdoor planar scenes from various positions resulting 1870 images in total. Sample images of the scenes are shown in Fig.~\ref{fig:dataset_scenes}. To obtain triplet correspondences we used SuperPoint features~\cite{detone2018superpoint} with LightGlue matches~\cite{lindenberger2023lightglue}. More details are available in SM.


For \case{1} we have randomly sampled 500 image triplets per camera per scene. For \case{2}/III we sampled 50 triplets par camera pair and for \case{4} 10 triplets per camera triplet. We only kept image triplets which had at least 10 triplet matches. For some scenes and cameras it was impossible to obtain 500, 50 or 10 triplets respectively resulting in 35 472 image triplets for \case{1}, 18 219 triplets for \case{2}/III and 12 876 for \case{4}. Details on the cameras, numbers of images and triplets are provided in SM.

\PAR{Evaluation.}
We evaluate the methods using $\xi_f$ \eqref{eqn:f_err} and the mean average accuracy derived from $\xi_f$~\cite{kocur2024robust}, which we denote as mAA$_f(t)$ representing the normalized area under the curve of the cumulative distribution function of $\xi_f$ on interval $\left[0, t\right]$.

Table~\ref{tab:eval} shows the results for our \hr{fff} and \hr{ff} solvers and the baselines for RANSAC 
with a fixed maximum number of 1000
iterations, with early termination at 0.9999 success probability  (allowed after the first 100 iterations), and epipolar threshold of 3px. The results show significantly better accuracy of \hr{fff} and \hr{ff} compared to the baseline methods.

To further demonstrate the efficacy of our method we also performed speed-vs-accuracy tradeoff evaluation by running each method for a different number of fixed RANSAC iterations. We measured runtime for all methods on one core of Intel Icelake 6338 2GHz processor. The results for our methods and the baselines are shown in Fig.~\ref{fig:eval_speed_accuracy}. 
Since our methods use the same non-linear optimization strategy as the baselines which utilize triplets, given enough iterations the different methods eventually converge to similar focal lengths resulting in similar accuracy. However, the results show that our solvers lead to very accurate results in fewer iterations compared to the baselines showcasing their practical viability for focal length estimation.


\PAR{Limitations.}\label{critical}
Our solvers have several limitations. First, the pure translation is a degenerate case for the $\M H_{fff}$ and $\M H_{f\rho\rho}$ solvers. However, if the focal length of the reference camera is known, we can still recover the focal lengths of the target cameras under pure translation.

Using the proposed solvers we may obtain multiple solutions. Therefore these solvers need to be used within a robust estimation framework such as RANSAC. Other simpler strategies such as voting~\cite{bujnak2009robust} may fail due to the fact that more than one solution may be geometrically feasible.

Moreover, as mentioned above for \case{3} and IV 
focal lengths can not be distinguished using only planar points thus requiring some off-plane points during scoring. Note that this is a property of the problem and not the solvers.
Problem with recovering one focal length for \case{3} was mentioned also in~\cite{heikkila2017using}.

%% file: sec/8_conclusion.tex
\section{Conclusion}

We address the problem of estimating the focal lengths of three cameras observing a planar scene. We derive novel constrains for this problem and use them to propose four new efficient solvers for different possible camera configurations. We extensively evaluate the proposed solvers on both real and synthetic data showing their superiority over baseline approaches. To the best of our knowledge, we are the first to perform such extensive evaluation for these problems. For this purpose, we introduce a new public dataset of planar scenes captured by multiple calibrated cameras.

\section*{Acknowledgments}

Y.D. and Z.K. were supported by the Czech Science Foundation (GAČR) JUNIOR STAR Grant (No. 22-23183M). 
V.K. and Z.B.H. were supported by the TERAIS project, a Horizon-Widera-2021 program of the European Union under the Grant agreement number 101079338 and by Slovak Research and Development Agency under project APVV-23-0250. Z. B. H. was supported by the grant KEGA 004UK-4/2024 “DICH: Digitalization of Cultural Heritage”. 
Q. W. and J. Y. were supported by the National Science Fund of China under Grant Nos. U24A20330 and 62361166670.
The results were obtained using the computational resources procured in the project National competence centre for high performance computing (project code:~311070AKF2) funded by European Regional Development Fund, EU Structural Funds Informatization of society, Operational Program Integrated Infrastructure.

%% file: sec/X_suppl.tex
\clearpage
\maketitlesupplementary

In this supplementary material, we provide additional information promised in the main paper. 
More detailed information on the constraints introduced in Sec.~4 of the main paper is provided in Sec.~\ref{sec:sm_constraints}. 
In Sec.~\ref{sec:sm_case4} we provide the details of the proposed \h{\rho f} solver for \case{4}. This solver was briefly introduced in Sec.~5.2 of the main paper. 
In Sec.~\ref{sec:sm_evaluation} we provide an evaluation of the proposed methods \hrf{\rho \rho f} and \hrf{\rho f} for \case{3} and IV using both synthetic and real data.
Sec.~\ref{sec:sm_dataset} contains information on the dataset that we have collected and used for evaluation in Sec.~6.2 of the main paper.

\section{New Constraints}

\label{sec:sm_constraints}

Here we provide details on the constraints introduced in Sec.~4 of the main paper. We show the steps for their derivation including Macaulay2~\cite{M2} code.

To derive the constraints relating the focal lengths and the elements of the matrix $\M Q_i$, we first create an ideal $I$~\cite{cox2005using} generated by the 12 polynomials extracted form the matrix equation Eq (16) in the main paper, \ie the equation
\begin{align}
\begin{split}
 [\M n]_\times \M Q_j [\M n]_\times^\top =s_j [\M n]_\times [\M n]_\times^\top,\ j=2,3 \\
 \end{split}
\label{eq:16_SM}
\end{align}
Note that both the left and right of~\eqref{eq:16_SM} are symmetric matrices, hence we can get 6 equations for each $j$. 

In the next step, the unknown elements of the normal vector $\M n$, \ie, $ n_x,n_y,n_z$, and the scale factors $s_2,s_3$ are eliminated from the generators of $I$ by computing the generators of the elimination ideal $J = I \cap \mathbb{C}[q_{21}, \dots , q_{36}]$. Here, $q_{j.}$ are the entries of $\M Q_2, \M Q_3$.
These generators can be computed, for example, in the computer algebra software Macaulay2~\cite{M2} using the following code:

{\scriptsize
\begin{verbatim}
KK = ZZ / 30097;
R = KK[q21,q22,q23,q24,q25,q26,
q31,q32,q33,q34,q35,q36,nx,ny,nz,s2,s3]
Q2 = matrix({{q21,q22,q23},{q22,q24,q25},{q23,q25,q26}});
Q3 = matrix({{q31,q32,q33},{q32,q34,q35},{q33,q35,q36}});
Nx = matrix {{0,-nz,ny},{nz,0,-nx},{-ny,nx,0}}; 
eqs = flatten(Nx*Q2*transpose(Nx)-s2*Nx*transpose(Nx) 
    | Nx*Q3*transpose(Nx)-s3*Nx*transpose(Nx) | nz-1) ;
I = ideal eqs;
J = eliminate(I,{nx,ny,nz,s2,s3});
\end{verbatim}
}

In this case, the elimination ideal $J$ is generated by seven polynomials $g_i$ of degree 6 in the elements of $\M Q_j$, $j =2,3$. The final constraints are only related to the 12 elements of the symmetric matrices $\M Q_i$ (6 from $\M Q_2$ and 6 from $\M Q_3$).  

\begin{figure*}[ht]
    \centering
\resizebox{\textwidth}{!}{
\begin{tikzpicture} 
    \begin{axis}[%
    hide axis, xmin=0,xmax=0,ymin=0,ymax=0,
    legend style={draw=white!15!white, 
    line width = 1pt,
    legend  columns =-1, 
    /tikz/every even column/.append style={column sep=0.2cm}
    }
    ]

    \addlegendimage{line width=4pt, Seaborn1}
    \addlegendentry{\hr{\rho ff} (ours)};
    \addlegendimage{line width=4pt, Seaborn2}
    \addlegendentry{\hr{\rho f} (ours)};
    \addlegendimage{line width=4pt, Seaborn3}
    \addlegendentry{\fEfrf};
    \addlegendimage{line width=4pt, Seaborn4}
    \addlegendentry{\fEfprf};
    \addlegendimage{line width=4pt, Seaborn5}
    \addlegendentry{\Efrf};    
    
    \end{axis}
    \end{tikzpicture}}

    \begin{subfigure}{0.31\textwidth}
    \includegraphics[width=\textwidth]{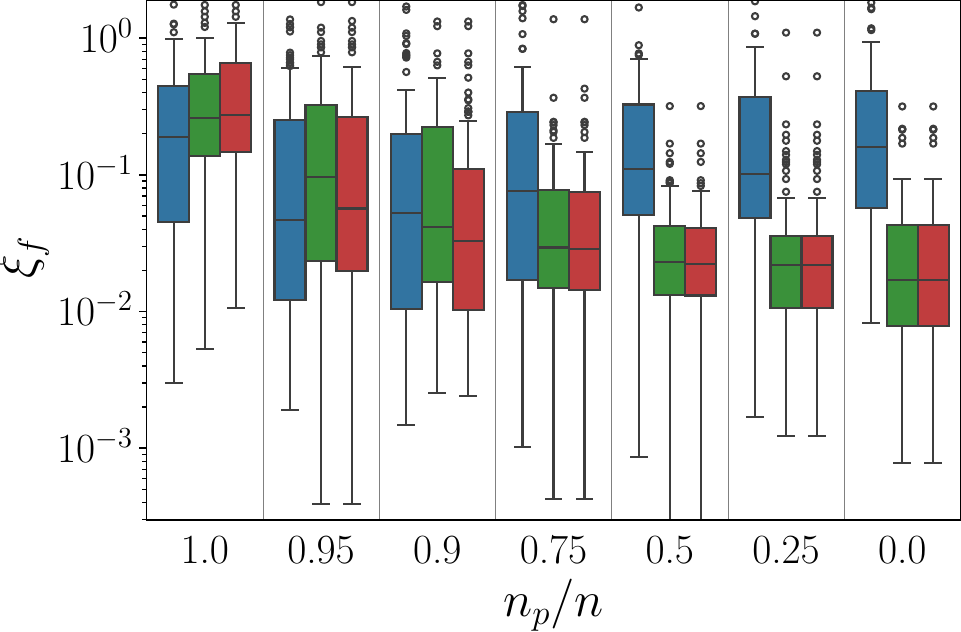}
    \caption{}
    \end{subfigure}
    \hfill    
    \begin{subfigure}{0.31\textwidth}
    \includegraphics[width=\textwidth]{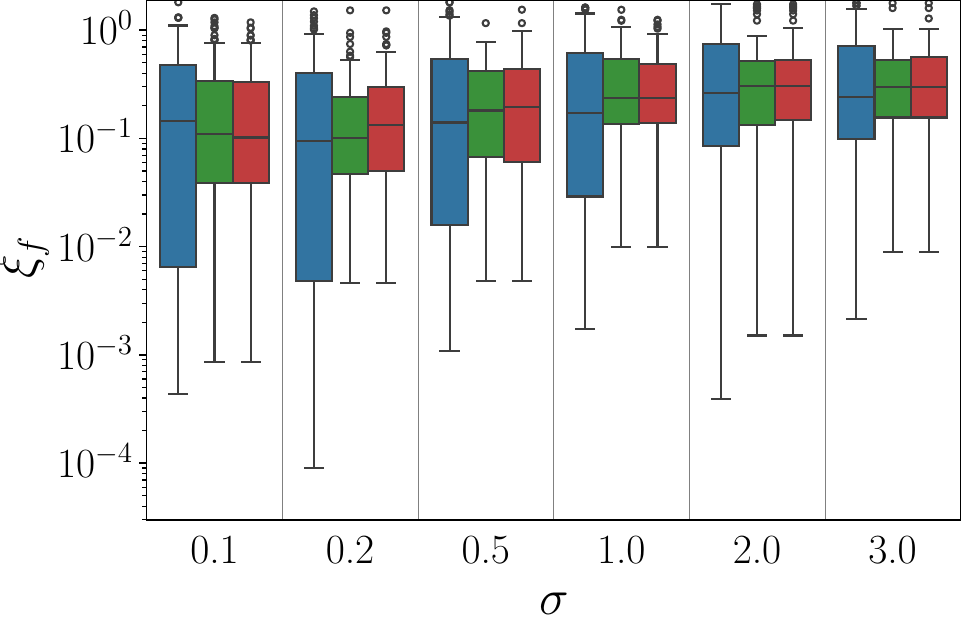}
    \caption{}
    \end{subfigure}
    \hfill    
    \begin{subfigure}{0.31\textwidth}
    \includegraphics[width=\textwidth]{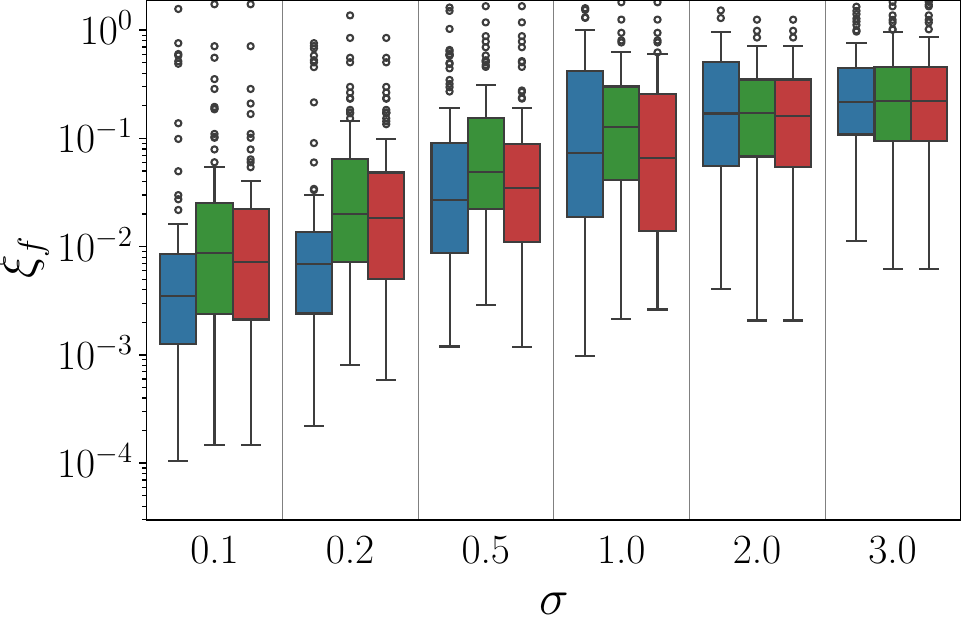}
    \caption{}
    \end{subfigure}

\vspace{1em}
    
    \begin{subfigure}{0.31\textwidth}
    \includegraphics[width=\textwidth]{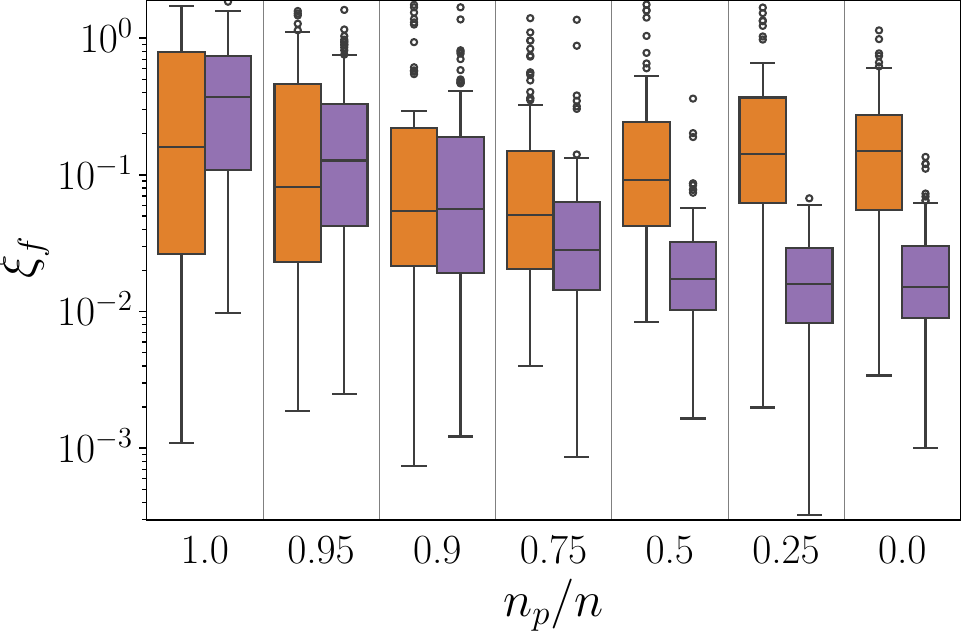}
    \caption{}
    \end{subfigure}
    \hfill    
    \begin{subfigure}{0.31\textwidth}
    \includegraphics[width=\textwidth]{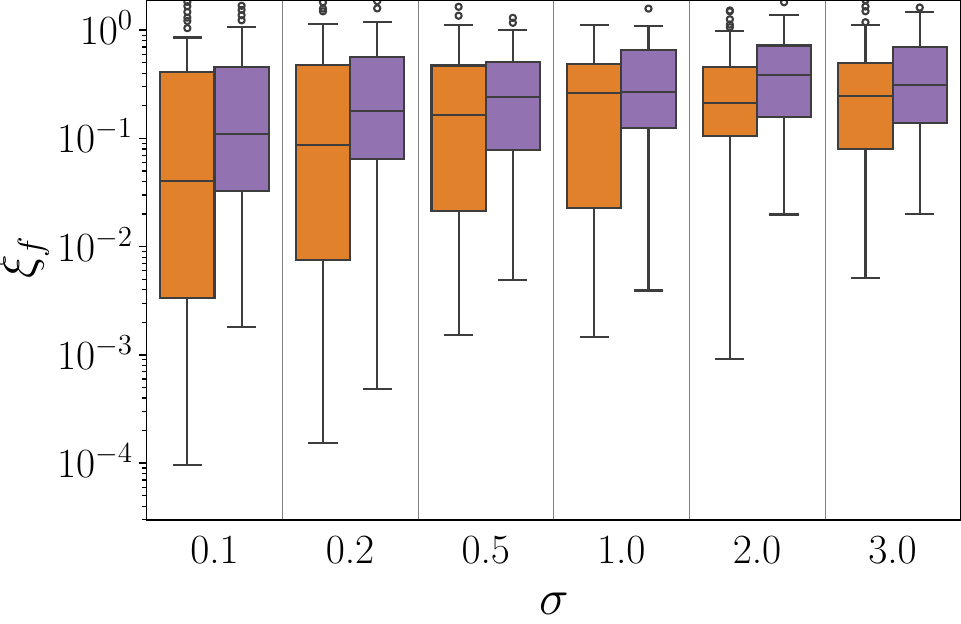}
    \caption{}
    \end{subfigure}
    \hfill    
    \begin{subfigure}{0.31\textwidth}
    \includegraphics[width=\textwidth]{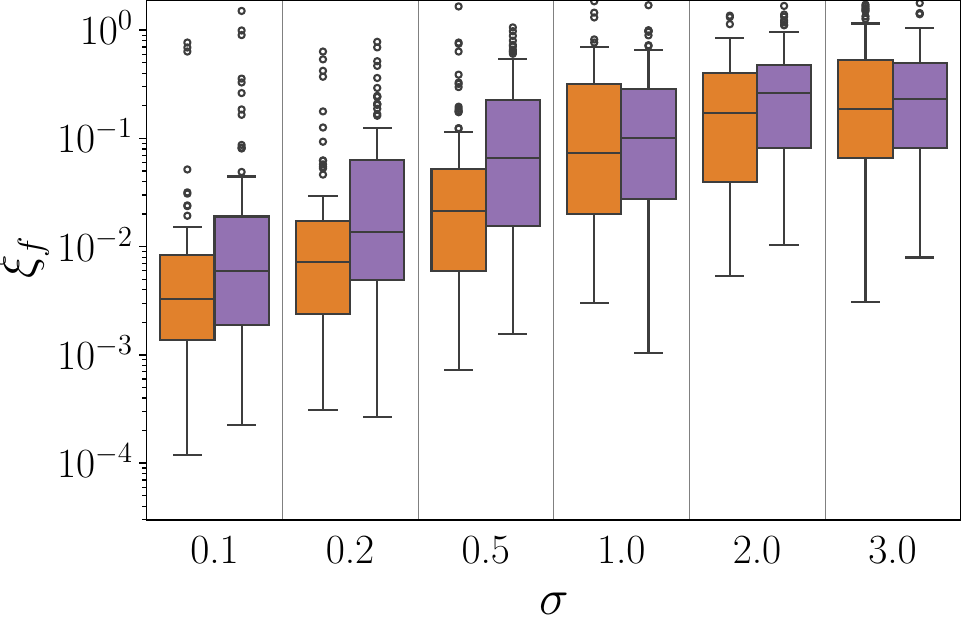}
    \caption{}
    \end{subfigure}

    \caption{Focal length errors for the evaluated methods in synthetic experiments with local optimization disabled. \textbf{\case{3}}: (a) We vary the proportion of points which lie on the dominant plane with fixed noise $\sigma = 1$. (b, c) We vary noise $\sigma$ with (b) $n_p / n = 1.0$ and (c) $n_p / n = 0.95$. \textbf{\case{4}}: (d,e,c) Same setup as for \case{3}. The synthetic setup is described in Sec.~6.1 if the main paper.}
    \label{fig:synth_accuracy_case34}
\end{figure*}

\section{Solver for Case \MakeUppercase{\romannumeral4}}

\label{sec:sm_case4}

In this section, we present details on the \h{\rho f} solver for \case{4}, which was introduced in Sec.~5.2 of the main paper. 

Similar to \case{3}, the system of polynomials $g_i$, $i=1,\dots,7$ can be written as ${\M M}{\M u} =\M 0$, where $\M M$ is a $7\times 16$ coefficient matrix and
\begin{equation}
{\M u}=[1,\beta,...,\beta^3,\alpha,\alpha\beta,...,\alpha^3,...,\alpha^3\beta^3]^\top,\label{eq:s25}
\end{equation}
is a vector consisting of the 16 monomials. We can  choose $\alpha$ as the hidden variable,
resulting in
\begin{equation}
{\M A}(\alpha){\tilde {\M u}} =\M 0, \label{eq:s26}
\end{equation}
where ${\M A}(\alpha)$ is a $7\times4$ polynomial matrix parameterized by $\alpha$, and ${\tilde {\M u}}=[1,\beta,...,\beta^3]^{\top}$ is a vector of 4 monomials in $\beta$ without $\alpha$. To solve the problem~\eqref{eq:s26} as a polynomial eigenvalue problem, it is sufficient to choose four out of the seven rows in~\eqref{eq:s26}  to get a square matrix $\tilde{\M A}(\alpha)$. In this case, we have  
\begin{equation}
\tilde{\M A}(\alpha) = \alpha^{3}{\M A}_{3} + \alpha^{2}{\M A}_{2} + \alpha{\M A}_1 + {\M A}_0, \label{eq:s27}
\end{equation}
where $\M A_3, \M A_2, \M A_1, \M A_0$ are $4\times 4$ matrices. The solutions to $\alpha$ are given by the eigenvalues of the following $12\times 12$ matrix 
\begin{equation}
    \begin{bmatrix}
\M 0 & {\M I} & \M 0 \\
\M 0 & \M 0 & {\M I}  \\
 -{\M A}_3^{-1}{\M A}_0 & -{\M A}_3^{-1}{\M A}_1 & -{\M A}_3^{-1}{\M A}_2
\end{bmatrix}\nonumber
\end{equation}

In this way, we obtain 12 possible solutions. The remaining steps are similar to \case{3}.  We denote this solver as \h{f \rho}.
Note that in this case, the original seven polynomial equations have only nine solutions. By selecting a subset of four polynomials, we introduced three more solutions. Still, the resulting solver is more efficient than the solver to the original seven equations, due to Gauss-Jordan elimination and computations of complex coefficients that are performed in a solver with 9 solutions.


\section{Evaluation for \case{3} and \case{4}}

\label{sec:sm_evaluation}

\begin{table*}
\resizebox{\linewidth}{!}{
\begin{tabular}{|l|rc|c|c|ccccc|} \cline{2-10}
\multicolumn{1}{c|}{} & \multicolumn{2}{|c|}{Method} & FOV Filtering & Sample & Median $\xi_f$ & Mean $\xi_f$ & mAA$_f$(0.1) & mAA$_f$(0.2) & Runtime (ms) \\ \hline
\multirow{ 9 }{*}{\rotatebox[origin=c]{90}{\case{ 3 }}}
 & \hr{f \rho \rho} & \textbf{ours} & \multirow{3}{*}{-} & 4 triplets & 0.2463&0.7266&19.34&29.68&121.62 \\
 & \fEfrf & & & 6 triplets & 0.2890&0.5889&16.33&25.84&\phantom{1}\textbf{34.09} \\
 & \fEfprf & & & 6 triplets & 0.2950&0.6253&16.04&25.46&\phantom{1}35.66 \\
 \cline{2-10} &  \hr{f \rho \rho} & \textbf{ours} & \multirow{3}{*}{35.5$^\circ$ - 61.5$^\circ$} & 4 triplets & 0.1478&0.5895&24.07&37.40&142.64 \\
 & \fEfrf & & & 6 triplets & 0.2136&0.4914&18.34&29.84&\phantom{1}40.57 \\
 & \fEfprf & & & 6 triplets & 0.2199&0.5127&17.89&29.23&\phantom{1}42.67 \\
 \cline{2-10} & \hr{f \rho \rho} & \textbf{ours} & \multirow{3}{*}{50$^\circ$ - 70$^\circ$} & 4 triplets & \textbf{0.1171}&0.4628&\textbf{27.22}&\textbf{42.32}&136.19 \\
 & \fEfrf & & & 6 triplets & 0.1582&\textbf{0.3744}&22.31&35.61&\phantom{1}37.33 \\
 & \fEfprf & & & 6 triplets & 0.1629&0.3781&21.95&35.09&\phantom{1}39.37 \\
\hline \hline
\multirow{ 6 }{*}{\rotatebox[origin=c]{90}{\case{ 4 }}}
 & \hr{f \rho} & \textbf{ours} & \multirow{2}{*}{-} & 4 triplets & 0.3286&0.5647&14.83&24.01&60.11 \\
 & \Efrf & & & 6 triplets & 0.3871&0.4881&11.99&19.61&\textbf{32.50} \\
 \cline{2-10} & \hr{f \rho} & \textbf{ours} & \multirow{2}{*}{35.5$^\circ$ - 61.5$^\circ$} & 4 triplets & 0.1559&0.2770&20.85&35.06&98.21 \\
 & \Efrf & &  & 6 triplets & 0.1982&0.3092&16.90&29.63&56.74 \\
 \cline{2-10} & \hr{f \rho} & \textbf{ours} &  \multirow{2}{*}{50$^\circ$ - 70$^\circ$} & 4 triplets & \textbf{0.1101}&\textbf{0.2258}&\textbf{26.52}&\textbf{43.30}&89.18 \\
 & \Efrf & & & 6 triplets & 0.1198&0.2342&24.84&41.36&50.77 \\
\hline
\end{tabular}}

\caption{Results on the real-world dataset of planar scenes for \case{3} and IV. FOV filtering denotes the range of FOV values that is used to reject models within RANSAC.}
\label{tab:SM_real}
\end{table*}

\begin{table*}
\resizebox{\linewidth}{!}{
\begin{tabular}{|l|c|rc|c|ccccc|} \cline{2-10}
\multicolumn{1}{c|}{} & Scene & \multicolumn{2}{|c|}{Method} & Sample & Median $\xi_f$ & Mean $\xi_f$ & mAA$_f$(0.1) & mAA$_f$(0.2) & Runtime (ms) \\ \hline

\multirow{ 6 }{*}{\rotatebox[origin=c]{90}{\case{ 3 }}}
 & \multirow{3}{*}{Original} & \hr{f \rho \rho} & \textbf{ours} & 4 triplets & 0.3076&1.1358&15.48&24.34&133.43 \\ 
 & & \fEfrf & & 6 triplets & 0.2494&0.5515&16.36&26.59&\phantom{1}\textbf{37.40} \\ 
 & & \fEfprf & & 6 triplets & 0.2620&0.5622&15.36&24.98&\phantom{1}39.44 \\ 
\cline{2-10}

& \multirow{3}{*}{Off-plane} & \hr{f \rho \rho} & \textbf{ours} & 4 triplets & \textbf{0.1540}&0.6324&\textbf{29.39}&\textbf{39.64}&142.92 \\ 
& & \fEfrf & & 6 triplets & 0.2035&\textbf{0.4716}&25.08&34.84&\phantom{1}38.36 \\ 
  & & \fEfprf & & 6 triplets & 0.2115&0.5959&24.60&34.30&\phantom{1}40.53 \\ 
 \hline
\hline
 
 \multirow{ 4 }{*}{\rotatebox[origin=c]{90}{\case{ 4 }}}
  & \multirow{2}{*}{Original} & \hr{f \rho} & \textbf{ours} & 4 triplets & 0.3843&0.5970&11.92&19.90&69.22 \\ 
 & & \Efrf & & 6 triplets & 0.4011&0.4896&11.00&18.98&\textbf{32.95} \\ \cline{2-10}
 & \multirow{2}{*}{Off-plane} 
  & \hr{f \rho} & \textbf{ours} &  4 triplets & \textbf{0.2381}&0.5239&\textbf{21.81}&\textbf{31.71}&69.97 \\ 
 & & \Efrf & & 6 triplets & 0.3150&\textbf{0.4300}&17.30&25.64&35.54 \\

\hline
\end{tabular}
}
\caption{Evaluation results on the original Book scene from the planar scenes dataset and its modified version with objects added in order to introduce off-plane scenes (see Fig.~\ref{fig:dinobook}). The modified version of the scene was captured by eight cameras. Therefore, for evaluation of the original scene we only consider triplets of images taken by the same eight cameras.}
\label{tab:offplane}
\end{table*}

In this section we evaluate the robust estimators \hr{\rho ff} for \case{3} and \hr{f \rho} for \case{4} which were presented in Sec.5.3 of the main paper. As baselines for comparison we use the combination of the \fEf solver~\cite{kukelova2012polynomial} or its DEGANSAC~\cite{chum2005degensac} variant with the plane and parallax solver~\cite{torii2011six} together with the \pnpf\xspace solver~\cite{kukelova2016efficient} for \case{3} and the \Ef\xspace solver in combination with \pnpf\xspace solver~\cite{kukelova2016efficient} for \case{4}.

We use the same evaluation metrics as in the main paper. Since two focal lengths are estimated jointly we use the geometric mean of their error $\xi_f = \sqrt{\xi_{f_1} \xi_{f_2}}$.

\subsection{Multiple Geometrically Valid Solutions}

We observe that for both \case{3} and \case{4} a single planar scene may result in multiple geometrically valid solutions. This makes it difficult to select a single correct solution for a given set of point correspondences. This leads to generally worse performance of the proposed solvers than for \case{1} and \case{2}. Note, that this is a feature of the problems and not the solvers.
Problem with recovering one focal length for \case{3} was mentioned also in~\cite{heikkila2017using}. 

In \case{3} and \case{4}, in contrast to baselines that use two-view \fEf and \Ef solvers and that completely fail for purely planar scenes, our solvers among the returned solutions contain the geometrically correct solution (see Figure 2 in the main paper). 
The proposed solvers just cannot distinguish between the returned solutions without additional information.
In Sec~\ref{sec:SM_real} we show how some simple strategies using prior knowledge about the focal lengths can significantly improve performance on real-world data even for these challenging problems.

We note that this problem can also be overcome whenever the scene contains a sufficient number of off-plane points. In such cases, there is one dominant plane with some off-plane objects visible in the three views. The off-plane points can then lead to higher scores for the correct solutions and are therefore selected during RANSAC. We demonstrate this both in synthetic experiments presented in the next section and with real-world experiments using which are presented in Sec.~\ref{sec:SM_real}.

\subsection{Synthetic Experiments}

We perform synthetic experiments with a setup similar to the one presented in Sec. 6.1 of the main paper. To better compare the performance of solvers under multiple possible valid solutions, we perform the experiment using vanilla RANSAC (without local optimization). The results of the synthetic experiments presented in Fig.~\ref{fig:synth_accuracy_case34} show that the estimators \hr{\rho ff} for \case{3} and \hr{f \rho} for \case{4} perform better than the baselines when considering a planar scene as well as scenes with a dominant plane. We also note that for all solvers the accuracy of the estimated focal lengths improves as more off-plane points are added to the planar scene.

\subsection{Real-World Experiments}

\begin{figure}[h]
\centering
    \begin{tabular}{cc}
    \includegraphics[width=0.45\linewidth]{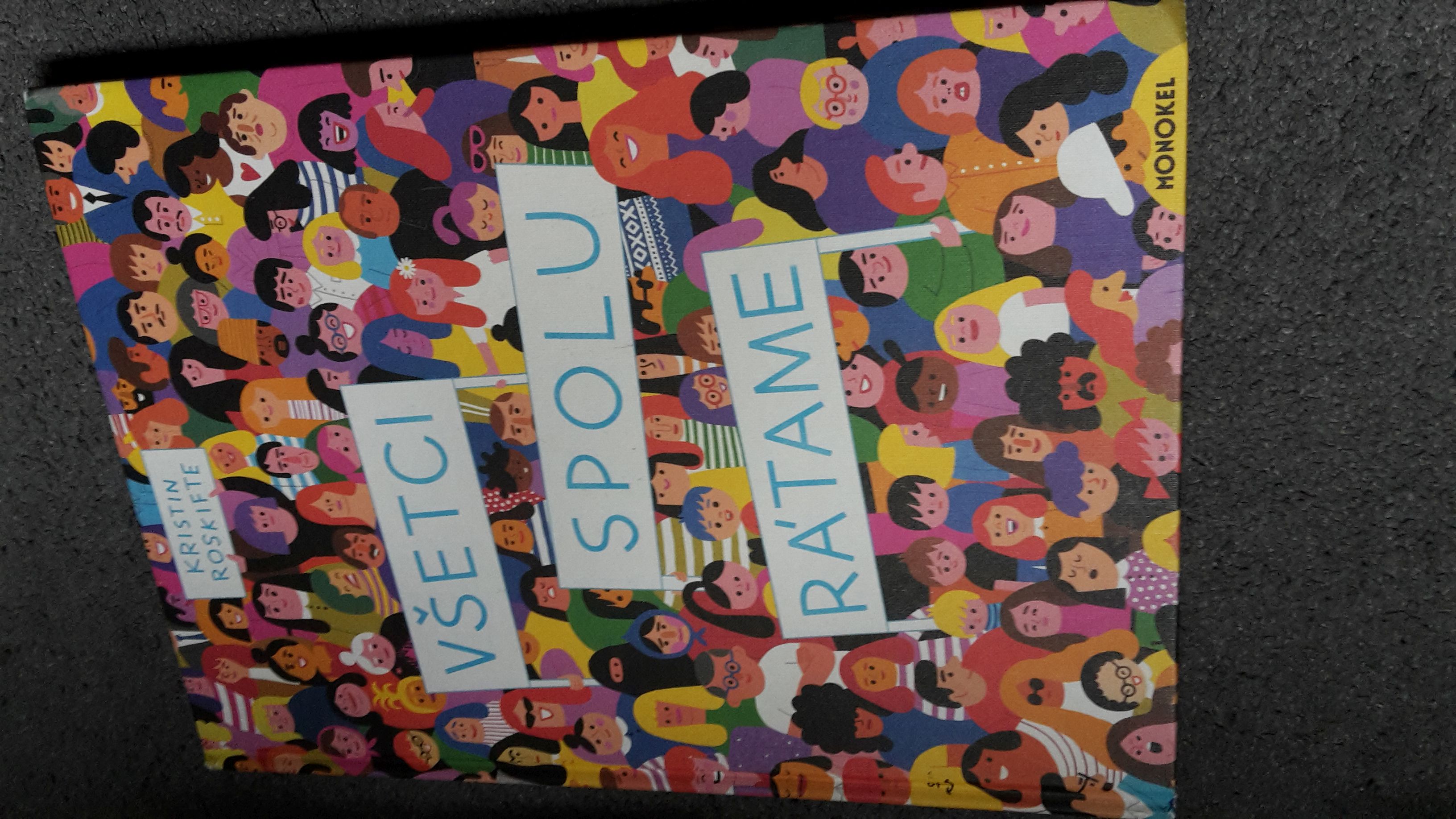} &
    \includegraphics[width=0.45\linewidth]{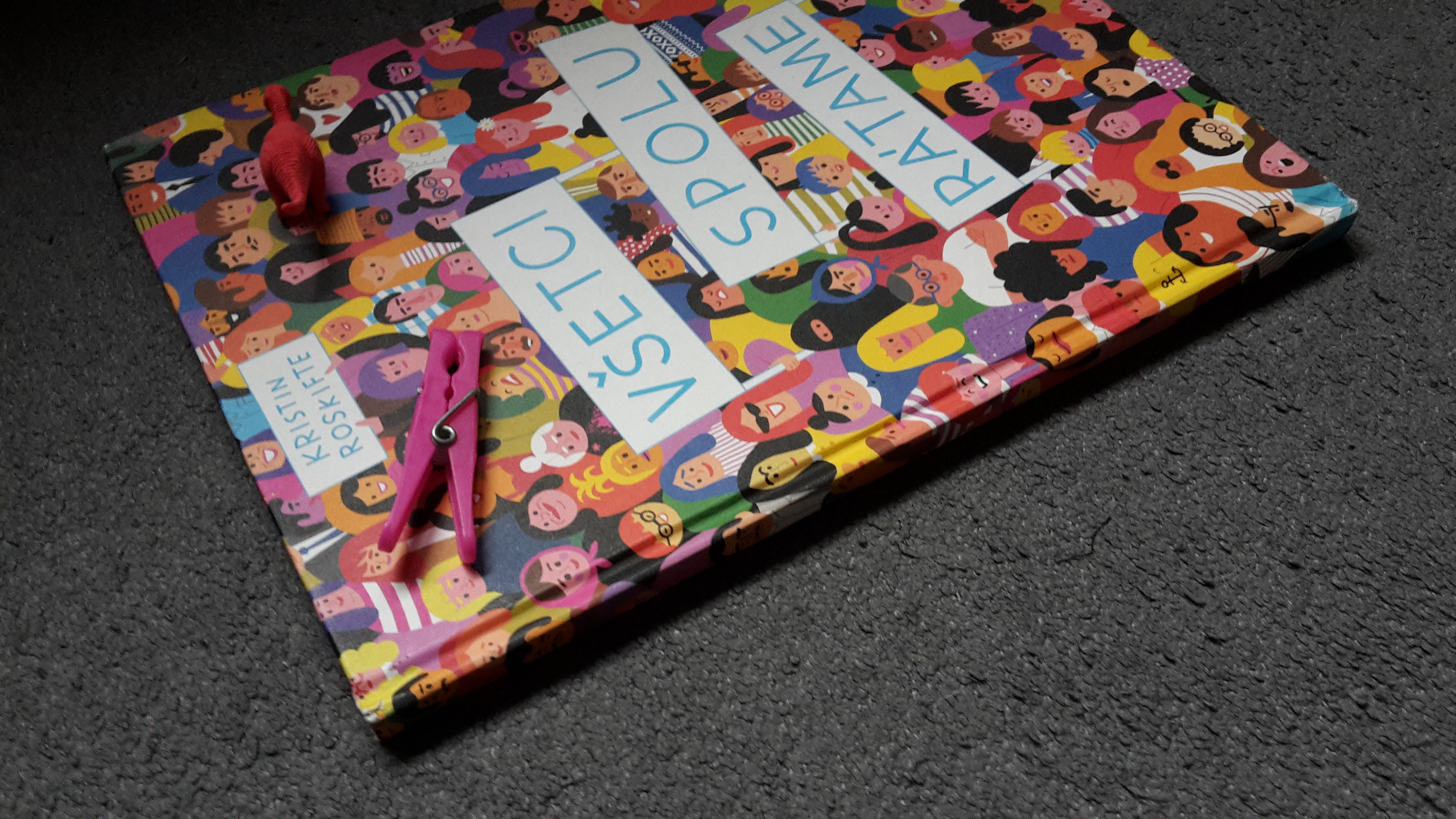} \\ 
    Original & Added off-plane objects
    \end{tabular}
    \caption{We added two objects to a scene from the planar dataset and recaptured it with 8 cameras that were also used to capture the original scene for comparison.}
    \label{fig:dinobook}
\end{figure}

\begin{table*}[ht]
\resizebox{0.99\linewidth}{!}{
\begin{tabular}{|c|c|ccc|ccccccc|ccc|}
\cline{6-15}
\multicolumn{5}{c}{} & \multicolumn{7}{|c|}{Images} & \multicolumn{3}{|c|}{Triplets} \\ \hline
ID & Description & FOV & Width & Height & Asphalt & Boats & Book & Facade & Floor & Papers & Calib & \case{1} & \case{2}/III & \case{4} \\ \hline
IPhoneOldBack &  Apple IPhone SE (2nd generation) back camera  & 62.8$^\circ$ & 4032 & 3024& 20&19&22&20&20&20&14 & 3000  & 2806& 2822 \\ \hline
IPhoneOldFront &  Apple IPhone SE (2nd generation) front camera  & 56.5$^\circ$ & 3088 & 2320& 20&18&18&23&17&21&18 & 2755  & 2984& 3089 \\ \hline
IPhoneZBHBack &  Apple IPhone SE (3rd generation) back camera  & 63.3$^\circ$ & 4032 & 3024& 19&19&21&20&20&20&20 & 3000  & 2761& 2781 \\ \hline
IPhoneZBHfront &  Apple IPhone SE (3rd generation) front camera  & 56.2$^\circ$ & 3088 & 2320& 20&20&22&19&21&20&20 & 2389  & 2900& 3065 \\ \hline
LenovoTabletBack & Tablet Lenovo TB-X505F back camera  & 59.7$^\circ$ & 2592 & 1944& 20&20&21&18&20&17&13 & 3000  & 2968& 3079 \\ \hline
LenovoTabletFront &  Tablet Lenovo TB-X505F front camera  & 62.3$^\circ$ & 1600 & 1200& \ding{55}&21&21&\ding{55}&16&\ding{55}&20 & 1500  & 1803& 2189 \\ \hline
MotoBack & Motorola Moto E4 Plus back camera  & 64.7$^\circ$ & 4160 & 3120& 20&23&22&21&22&20&37 & 2013  & 2529& 2903 \\ \hline
MotoFront & Motorola Moto E4 Plus front camera  & 71.0$^\circ$ & 2592 & 1952& \ding{55}&18&19&\ding{55}&\ding{55}&\ding{55}&19 & 1000  & 1400& 1732 \\ \hline
Olympus & Olympus uD600,S600 compact digital camera  & 49.1$^\circ$ & 2816 & 2112& 19&23&24&17&21&21&23 & 3000  & 3145& 3171 \\ \hline
SamsungBack & Samsung Galaxy S5 Mini back camera  & 56.1$^\circ$ & 3264 & 1836& \ding{55}&20&28&\ding{55}&20&20&18 & 2000  & 2356& 2594 \\ \hline
SamsungFront & Samsung Galaxy S5 Mini front camera  & 69.4$^\circ$ & 1920 & 1080& 19&24&19&23&19&20&20 & 2798  & 3023& 3120 \\ \hline
SamsungGlossyBack & Samsung Galaxy S III Mini back camera  & 53.8$^\circ$ & 2560 & 1920& 20&21&19&20&19&20&21 & 3000  & 3073& 3180 \\ \hline
SamsungGlossyFront & Samsung Galaxy S III Mini front camera  & 55.6$^\circ$ & 640 & 480& 21&20&26&20&20&21&20 & 3000  & 1790& 1726 \\ \hline
DellWide & Dell Precision 7650 notebook camera  & 80.0$^\circ$ & 1280 & 720& \ding{55}&21&22&\ding{55}&22&\ding{55}&20 & 1500  & 1254& 1446 \\ \hline
SonyTelescopic & Sony $\alpha$5000 digital camera with 55-210mm Lens  & 23.5$^\circ$ & 5456 & 3064& 20&20&20&22&18&\ding{55}&20 & 1517  & 1664& 1815 \\ \hline
\multicolumn{2}{c}{} & \multicolumn{3}{|c|}{Total} & 218&307&324&223&275&220&303 &35472 &18219 & 12876 \\ \cline{3-15}
\multicolumn{2}{c}{} & \multicolumn{3}{|c|}{Triplets \case{1}} & 3574&7500&7500&5500&5898&5500& \ding{55} & \multicolumn{2}{c}{} \\ \cline{3-12}
\multicolumn{2}{c}{} & \multicolumn{3}{|c|}{Triplets \case{2}} & 649&5100&4856&2538&2555&2521& \ding{55} & \multicolumn{2}{c}{} \\ \cline{3-12}
\multicolumn{2}{c}{} & \multicolumn{3}{|c|}{Triplets \case{4}} & 252&4256&3871&1472&1618&1407& \ding{55} & \multicolumn{2}{c}{} \\ \cline{3-12}
\end{tabular}}
\caption{Summary of our evaluation dataset. The table shows the number of included images per scene per camera and the number of extracted triplets. The last three columns indicate how many triplets for a given case contain an image from a given camera (\eg for \case{4} we use 1815 triplets for which at least one of the images was taken using Dell Precision 7650 notebook camera). The fourth row from bottom denotes the total number of images per scene in the dataset and and in the last three columns the total number of triplets per case. 
The last three rows show how many triplets are included for each scene.
}
\label{tab:dataset}
\end{table*}

\label{sec:SM_real}

We also perform experiments on the real-world dataset introduced in this paper. To overcome issues with multiple geometrically feasible solutions, we propose a simple strategy of  acceptable field-of-view (FOV) ranges for the focal lengths. During RANSAC we simply discard all solutions with focal lengths outside of the predetermined range.

In Tab.~\ref{tab:SM_real} we show the results comparing our method and the baseline approaches in three different variants. As the first variant we do not discard any solutions. For the second variant, we set range of acceptable FOVs by considering the prior for focal lengths used by the popular SfM software COLMAP~\cite{schonberger2016structure} which is set as $f_p = 1.2\text{max}(width, height)$, which corresponds to a field of view of $\sim 45^\circ$. To obtain a range we consider 30\% increase or decrease in focal length, resulting in the acceptable field of view range from 35.5$^\circ$ to 61.5$^\circ$. As the last variant we use a range of 50$^\circ$ to 70$^\circ$. We chose this range since most commercially available phone cameras fall within it.

All variants were implemented in PoseLib~\cite{poselib}. We set the maximum epipolar threshold to 3 px. We used early termination with 0.9999 confidence and a minimum of 100 iterations and a maximum of 1000.

For all of the proposed variants and both cases our methods \hr{\rho ff} and \hr{\rho f} show superior performance in terms of the accuracy of the estimated focal lengths compared to the baselines. We also note that the performance of all evaluated methods significantly improves when filtering solutions based on the predetermined field of view ranges. Showcasing how a simple strategy can significantly improve the accuracy of all evaluated methods.

\PAR{Evaluation using a scene with off-plane points}

We also perform additional evaluation with one scene which is similar to the Book scene from our planar dataset, but we include additional objects (see Fig.~\ref{fig:dinobook}) to introduce some off-plane points. This sequence was captured by 8 of the cameras used to capture the planar dataset. It contains 87 images in total and we used them to generate 5066 triplets for \case{3} and 1473 triplets for \case{4}.\footnote{Note that this scene is not included in the dataset presented in Sec.~\ref{sec:sm_dataset} due to its different nature, \ie containing additional non-planar objects.} We perform a comparison between the results obtained on this scene with off-plane objects and results obtained on the original planar scene. For the original planar scene we only consider triplets using the same cameras as in the  scene with off-plane objects. Tab.~\ref{tab:offplane} shows a comparison of the results for all evaluated methods. The results show a significant improvement of all methods when off-plane objects are introduced in the scene. This shows that the problem becomes easier when the scene is not fully planar, but retains a significant dominant plane with some off-plane points. In this scenario, our methods show significantly better performance over the baselines.

\section{Real World Dataset}

\label{sec:sm_dataset}


This section provides the detailed information about the dataset used for real-world evaluation presented in Sec.~6.2 of the main paper. The descriptions of the cameras and dataset statistics regarding the total number of images and extracted triplets are provided in Tab.~\ref{tab:dataset}. The dataset contains 1870 images of  4 indoor and 2 outdoor planar scenes captured with 14 calibrated cameras. We purposefully select some scenes to be more challenging (\eg repeating patterns in Floor, few significant landmarks in Asphalt). In total we use provide 66 567 image triplets for evaluation of the different cases.

\subsection{Calibration}

To calibrate the cameras, we used a standard checkerboard pattern printed on hard plastic. We manually removed blurry or otherwise unsuitable calibration images from the dataset. We calibrated the cameras using~\cite{zhang2000calibration}. During calibration, we used the assumption of square pixels (\ie $f_x = f_y$). We also modeled tangential and radial distortion to obtain more accurate focal lengths. All used cameras exhibited low distortion so we use the original distorted images for evaluation to better reflect accuracy in real scenarios where cameras are expected to have low and unknown distortion. The images used for calibration, as well as the calibration code and estimated intrinsics, will be made available with the dataset.

\subsection{Triplet Point Correspondence Extraction}

To obtain triplet correspondences we used SuperPoint~\cite{detone2018superpoint} with inference in the original image resolution keeping at most 2048 best keypoints. We matched the keypoints using LightGlue~\cite{lindenberger2023lightglue} to first perform pairwise matches. Since SuperPoint was trained to only be rotationally invariant up to $45^\circ$ rotations we have extracted pairwise matches by rotating one of the images four times with a step of $90^\circ$ and selecting the orientation which produced the largest number of matches. Afterwards, we kept only those correspondences which were matched across all three pairs thus producing triplets. We will provide the extracted correspondences as part of the dataset upon release.

%% file: main.bib
@string{pami = "Trans. Pattern Analysis and Machine Intelligence (PAMI)"}

@string{eccv = "European Conference on Computer Vision (ECCV)"}

@string{iccv = "International Conference on Computer Vision (ICCV)"}

@string{accv = "Asian Conference on Computer Vision (ACCV)"}

@string{cvpr = "Computer Vision and Pattern Recognition (CVPR)"}

@book{hartley2003multiple,
  title={Multiple view geometry in computer vision},
  author={Hartley, Richard and Zisserman, Andrew},
  year={2003},
  publisher={Cambridge university press}
}

@inproceedings{schonberger2016structure,
  title={Structure-from-motion revisited},
  author={Schonberger, Johannes L and Frahm, Jan-Michael},
  booktitle=cvpr,
  pages={4104--4113},
  year={2016}
}

@inproceedings{ding2023minimal,
  title={Minimal Solutions to Generalized Three-View Relative Pose Problem},
  author={Ding, Yaqing and Chien, Chiang-Heng and Larsson, Viktor and {\AA}str{\"o}m, Karl and Kimia, Benjamin},
  booktitle=iccv,
  year={2023}
}

@inproceedings{ding2023revisiting,
  title={Revisiting the P3P problem},
  author={Ding, Yaqing and Yang, Jian and Larsson, Viktor and Olsson, Carl and {\AA}str{\"o}m, Kalle},
  booktitle=cvpr,
  year={2023}
}

@book{maybank1993theory,
  title={Theory of reconstruction from image motion},
  author={Maybank, Stephen},
  year={1993},
  publisher={Springer}
}

@inproceedings{heikkila2017using,
  title={Using sparse elimination for solving minimal problems in computer vision},
  author={Heikkilä, Janne},
  booktitle={Proceedings of the IEEE International Conference on Computer Vision},
  year={2017}
}

@article{malis2002camera,
  title={Camera self-calibration from unknown planar structures enforcing the multiview constraints between collineations},
  author={Malis, Ezio and Cipolla, Roberto},
  journal={IEEE Transactions on Pattern Analysis and Machine Intelligence},
  year={2002},
  publisher={IEEE}
}

@Misc{M2,
  author = {Grayson, Daniel R. and Stillman, Michael E.},
  title = {Macaulay2, a software system for research in algebraic geometry},
  howpublished = {Available at \url{http://www2.macaulay2.com}}
}

@book{cox2005using,
  title={Using algebraic geometry},
  author={Cox, David A and Little, John and O'shea, Donal},
  year={2005},
  publisher={Springer Science \& Business Media}
}

@inproceedings{kukelova2017clever,
    title={A clever elimination strategy for efficient minimal solvers},
    author={Kukelova, Zuzana and Kileel, Joe and Sturmfels, Bernd and Pajdla, Tomas},
    booktitle=cvpr,
    year={2017}
}

@article{hartley2012efficient,
	title={An efficient hidden variable approach to minimal-case camera motion estimation},
	author={Hartley, Richard and Li, Hongdong},
	journal=pami,
	year={2012}
}

@inproceedings{larsson2017efficient,
	title={Efficient Solvers for Minimal Problems by Syzygy-Based Reduction.},
	author={Larsson, Viktor and {\AA}str{\"o}m, Kalle and Oskarsson, Magnus},
	booktitle=cvpr,
	year={2017}
}

@book{gellert2012vnr,
  title={The VNR concise encyclopedia of mathematics},
  author={Gellert, Walter and Hellwich, M. and K{\"a}stner, H and K{\"u}stner, H},
  year={2012},
  publisher={Springer Science \& Business Media}
}

@inproceedings{larsson2018beyond,
	title={Beyond Gr{\"o}Bner Bases: Basis Selection for Minimal Solvers},
	author={Larsson, Viktor and Oskarsson, Magnus and {\AA}str{\"o}m, Kalle and Wallis, Alge and Kukelova, Zuzana and Pajdla, Tomas},
	booktitle=cvpr,
	year={2018}
}

@InProceedings{Bhayani_2020_CVPR,
author = {Bhayani, Snehal and Kukelova, Zuzana and Heikkila, Janne},
title = {A Sparse Resultant Based Method for Efficient Minimal Solvers},
booktitle = cvpr,
year = {2020}
}

@article{kukelova2012polynomial,
	title={Polynomial eigenvalue solutions to minimal problems in computer vision},
	author={Kukelova, Zuzana and Bujnak, Martin and Pajdla, Tomas},
	journal=pami,
	year={2012}
}

@book{bai2000templates,
	title={Templates for the solution of algebraic eigenvalue problems: a practical guide},
	author={Bai, Zhaojun and Demmel, James and Dongarra, Jack and Ruhe, Axel and van der Vorst, Henk},
	year={2000},
	publisher={SIAM}
}

@article{stefanovic1973relative,
  title={Relative orientation--a new approach},
  author={Stefanovic, P},
  journal={ITC Journal},
  volume={3},
  number={417-448},
  pages={2},
  year={1973}
}

@article{nister2004efficient,
  title={An efficient solution to the five-point relative pose problem},
  author={Nist{\'e}r, David},
  journal={IEEE transactions on pattern analysis and machine intelligence},
  year={2004},
  publisher={IEEE}
}

@article{ding2020homography,
  title={Homography-based minimal-case relative pose estimation with known gravity direction},
  author={Ding, Yaqing and Yang, Jian and Ponce, Jean and Kong, Hui},
  journal={IEEE transactions on pattern analysis and machine intelligence},
  year={2022},
  publisher={IEEE}
}

@article{zhang2000calibration,
  title={A flexible new technique for camera calibration},
  author={Zhang, Zhengyou},
  journal={IEEE Transactions on pattern analysis and machine intelligence},
  volume={22},
  number={11},
  pages={1330--1334},
  year={2000},
  publisher={IEEE}
}

@inproceedings{detone2018superpoint,
  title={Superpoint: Self-supervised interest point detection and description},
  author={DeTone, Daniel and Malisiewicz, Tomasz and Rabinovich, Andrew},
  booktitle={Proceedings of the IEEE conference on computer vision and pattern recognition workshops},
  pages={224--236},
  year={2018}
}

@inproceedings{lindenberger2023lightglue,
  title={Lightglue: Local feature matching at light speed},
  author={Lindenberger, Philipp and Sarlin, Paul-Edouard and Pollefeys, Marc},
  booktitle={Proceedings of the IEEE/CVF International Conference on Computer Vision},
  pages={17627--17638},
  year={2023}
}

@inproceedings{chum2003loransac,
  title={Locally optimized RANSAC},
  author={Chum, Ond{\v{r}}ej and Matas, Ji{\v{r}}{\'\i} and Kittler, Josef},
  booktitle={Pattern Recognition: 25th DAGM Symposium, Magdeburg, Germany, September 10-12, 2003. Proceedings 25},
  pages={236--243},
  year={2003},
  organization={Springer}
}

@misc{poselib,
  title = {{PoseLib - Minimal Solvers for Camera Pose Estimation}},
  author = {Viktor Larsson and contributors},
  URL = {https://github.com/vlarsson/PoseLib},
  year = {2020}
}

@inproceedings{kocur2024robust,
  title={Robust Self-calibration of Focal Lengths from the Fundamental Matrix},
  author={Kocur, Viktor and Kyselica, Daniel and Kukelova, Zuzana},
  booktitle={Proceedings of the IEEE/CVF Conference on Computer Vision and Pattern Recognition},
  pages={5220--5229},
  year={2024}
}

@inproceedings{cin2024minimal,
  title={Minimal Perspective Autocalibration},
  author={Cin, Andrea Porfiri Dal and Duff, Timothy and Magri, Luca and Pajdla, Tomas},
  booktitle=cvpr,
  year={2024}
}

@inproceedings{torii2011six,
  title={The six point algorithm revisited},
  author={Torii, Akihiko and Kukelova, Zuzana and Bujnak, Martin and Pajdla, Tomas},
  booktitle={Computer Vision--ACCV 2010 Workshops: ACCV 2010 International Workshops, Queenstown, New Zealand, November 8-9, 2010, Revised Selected Papers, Part II 10},
  pages={184--193},
  year={2011},
  organization={Springer}
}

@inproceedings{bougnoux1998projective,
  title={From projective to euclidean space under any practical situation, a criticism of self-calibration},
  author={Bougnoux, Sylvain},
  booktitle={Sixth International Conference on Computer Vision},
  pages={790--796},
  year={1998},
  organization={IEEE}
}

@article{tzamos2023relative,
  title={Relative pose of three calibrated and partially calibrated cameras from four points using virtual correspondences},
  author={Tzamos, Charalambos and Barath, Daniel and Sattler, Torsten and Kukelova, Zuzana},
  journal={arXiv preprint arXiv:2303.16078},
  year={2023}
}

@inproceedings{chum2005degensac,
  title={Two-view geometry estimation unaffected by a dominant plane},
  author={Chum, Ondrej and Werner, Tomas and Matas, Jiri},
  booktitle={2005 IEEE Computer Society Conference on Computer Vision and Pattern Recognition (CVPR'05)},
  volume={1},
  pages={772--779},
  year={2005},
  organization={IEEE}
}

@inproceedings{bujnak20093d,
  title={3d reconstruction from image collections with a single known focal length},
  author={Bujnak, Martin and Kukelova, Zuzana and Pajdla, Tomas},
  booktitle={2009 IEEE 12th International Conference on Computer Vision},
  pages={1803--1810},
  year={2009},
  organization={IEEE}
}

@inproceedings{kukelova2008automatic,
  title={Automatic generator of minimal problem solvers},
  author={Kukelova, Zuzana and Bujnak, Martin and Pajdla, Tomas},
  booktitle={Computer Vision--ECCV 2008: 10th European Conference on Computer Vision, Marseille, France, October 12-18, 2008, Proceedings, Part III 10},
  pages={302--315},
  year={2008},
  organization={Springer}
}

@inproceedings{kukelova2016efficient,
  title={Efficient intersection of three quadrics and applications in computer vision},
  author={Kukelova, Zuzana and Heller, Jan and Fitzgibbon, Andrew},
  booktitle={Proceedings of the IEEE Conference on Computer Vision and Pattern Recognition},
  pages={1799--1808},
  year={2016}
}

@article{nister2006Four,
  title = {Four {{Points}} in {{Two}} or {{Three Calibrated Views}}: {{Theory}} and {{Practice}}},
  shorttitle = {Four {{Points}} in {{Two}} or {{Three Calibrated Views}}},
  author = {Nist{\'e}r, David and Schaffalitzky, Frederik},
  year = {2006},
  journal = {International Journal of Computer Vision}
}

@article{quan2006Results,
  title = {Some {{Results}} on {{Minimal Euclidean Reconstruction}} from {{Four Points}}},
  author = {Quan, Long and Triggs, Bill and Mourrain, Bernard},
  year = {2006},
  journal = {Journal of Mathematical Imaging and Vision}
}

@inproceedings{hruby2022Learning,
  title = {Learning to {{Solve Hard Minimal Problems}}},
  booktitle = cvpr,
  author = {Hruby, Petr and Duff, Timothy and Leykin, Anton and Pajdla, Tomas},
  year = {2022}
}

@inproceedings{hartley1992estimation,
  title={Estimation of relative camera positions for uncalibrated cameras},
  author={Hartley, Richard},
  booktitle=eccv,
  year={1992},
  organization={Springer}
}

@book{sommese2005numerical,
  title={The Numerical solution of systems of polynomials arising in engineering and science},
  author={Sommese, Andrew J and Wampler, Charles W},
  year={2005},
  publisher={World Scientific}
}

@article{hauenstein2018adaptive,
  title={Adaptive strategies for solving parameterized systems using homotopy continuation},
  author={Hauenstein, Jonathan D and Regan, Margaret H},
  journal={Applied Mathematics and Computation},
  year={2018},
  publisher={Elsevier}
}

@inproceedings{chien2022gpu,
  title={{GPU}-based homotopy continuation for minimal problems in computer vision},
  author={Chien, Chiang-Heng and Fan, Hongyi and Abdelfattah, Ahmad and Tsigaridas, Elias and Tomov, Stanimire and Kimia, Benjamin},
  booktitle=cvpr,
  year={2022}
}

@inproceedings{stewenius2005minimal,
	title={A minimal solution for relative pose with unknown focal length},
	author={Stew{\'e}nius, Henrik and Nist{\'e}r, David and Kahl, Fredrik and Schaffalitzky, Frederik},
	booktitle=cvpr,
	year={2005}
}

@inproceedings{brown2007minimal,
  title={Minimal solutions for panoramic stitching},
  author={Brown, Matthew and Hartley, Richard I and Nist{\'e}r, David},
  booktitle=cvpr,
  year={2007}
}

@inproceedings{li2006simple,
	title={A simple solution to the six-point two-view focal-length problem},
	author={Li, Hongdong},
	booktitle=eccv,
	year={2006},
	organization={Springer}
}

@article{hartley1997defense,
  title={In defense of the eight-point algorithm},
  author={Hartley, Richard I},
  journal={IEEE Transactions on pattern analysis and machine intelligence},
  year={1997}
}

@inproceedings{bujnak2009robust,
  title={Robust focal length estimation by voting in multi-view scene reconstruction},
  author={Bujnak, Martin and Kukelova, Zuzana and Pajdla, Tomas},
  booktitle={Asian Conference on Computer Vision},
  pages={13--24},
  year={2009},
  organization={Springer}
}
